\PassOptionsToPackage{table}{xcolor} 
\documentclass[sigconf,9pt]{acmart}
\AtBeginDocument{%
  }

\setcopyright{acmlicensed}
\copyrightyear{2026}
\acmYear{2026}
\acmDOI{XXXXXXX.XXXXXXX}
\acmConference[SenSys '26]{ACM/IEEE International Conference on Embedded Artificial Intelligence and Sensing Systems}{May 11--14,
  2026}{Saint-Malo, France}
\acmISBN{978-1-4503-XXXX-X/2026/05}

\usepackage{graphicx}
\usepackage{multirow}
\usepackage{hhline}
\usepackage{pifont}
\usepackage{enumitem}
\usepackage{upgreek}
\usepackage{subcaption}

\usepackage{amssymb}
\newcommand{\xmark}{\ding{55}}

\begin{document}

\title[NeRC: Neural Ranging Correction through Differentiable MHLE]{NeRC: Neural Ranging Correction through \\ Differentiable Moving Horizon Location Estimation}

\author{Xu Weng}
\authornote{Co-corresponding authors.}
\orcid{0000-0003-3976-4705}
\affiliation{%
  \institution{Nanyang Technological University}
  \city{Singapore}
  \country{Singapore}
}
\email{xu009@e.ntu.edu.sg}

\author{K.V. Ling}
\orcid{0000-0002-9293-9394}
\affiliation{%
  \institution{Nanyang Technological University}
  \city{Singapore}
  \country{Singapore}}
\email{ekvling@ntu.edu.sg}

\author{Haochen Liu}
\orcid{0000-0002-3628-8777}
\affiliation{%
 \institution{Nanyang Technological University}
 \city{Singapore}
 \country{Singapore}}
\email{haochen002@e.ntu.edu.sg}

\author{Bingheng Wang}
\orcid{0000-0002-1040-8515}
\authornotemark[1]
\affiliation{%
  \institution{National University of Singapore}
  \city{Singapore}
  \country{Singapore}}
\email{wangbingheng@u.nus.edu}

\author{Kun Cao}
\orcid{0000-0003-4688-1096}
\affiliation{%
  \institution{Tongji University}
  \city{Shanghai}
  \country{China}
}
\additionalaffiliation{
\institution{the Shanghai Institute of Intelligent Science and Technology, National Key Laboratory of Autonomous Intelligent Unmanned Systems, and Frontiers Science Center for Intelligent Autonomous Systems, Ministry of Education, Beijing, China}
}
\email{caokun@tongji.edu.cn}

\renewcommand{\shortauthors}{Xu Weng et al.}

\begin{abstract}
  GNSS localization using everyday mobile devices is challenging in urban environments, as ranging errors caused by the complex propagation of satellite signals and low-quality onboard GNSS hardware are blamed for undermining positioning accuracy. Researchers have pinned their hopes on data-driven methods to regress such ranging errors from raw measurements. However, the grueling annotation of ranging errors impedes their pace. This paper presents a robust end-to-end Neural Ranging Correction (NeRC) framework, where localization-related metrics serve as the task objective for training the neural modules. Instead of seeking impractical ranging error labels, we train the neural network using ground-truth locations that are relatively easy to obtain. This functionality is supported by differentiable moving horizon location estimation (MHE) that handles a horizon of measurements for positioning and backpropagates the gradients for training. Even better, as a blessing of end-to-end learning, we propose a new training paradigm using Euclidean Distance Field (EDF) cost maps, which further alleviates the demands on labeled locations. We evaluate NeRC on public benchmarks and our collected datasets, demonstrating its distinguished improvement in positioning accuracy. We also deploy NeRC on the edge to verify its real-time performance for mobile devices.
\end{abstract}




\received{20 February 2007}
\received[revised]{12 March 2009}
\received[accepted]{5 June 2009}

\maketitle

\section{Introduction}
\begin{figure}[!t]
  \centering
\subfloat[]{\includegraphics[width=0.5\linewidth]{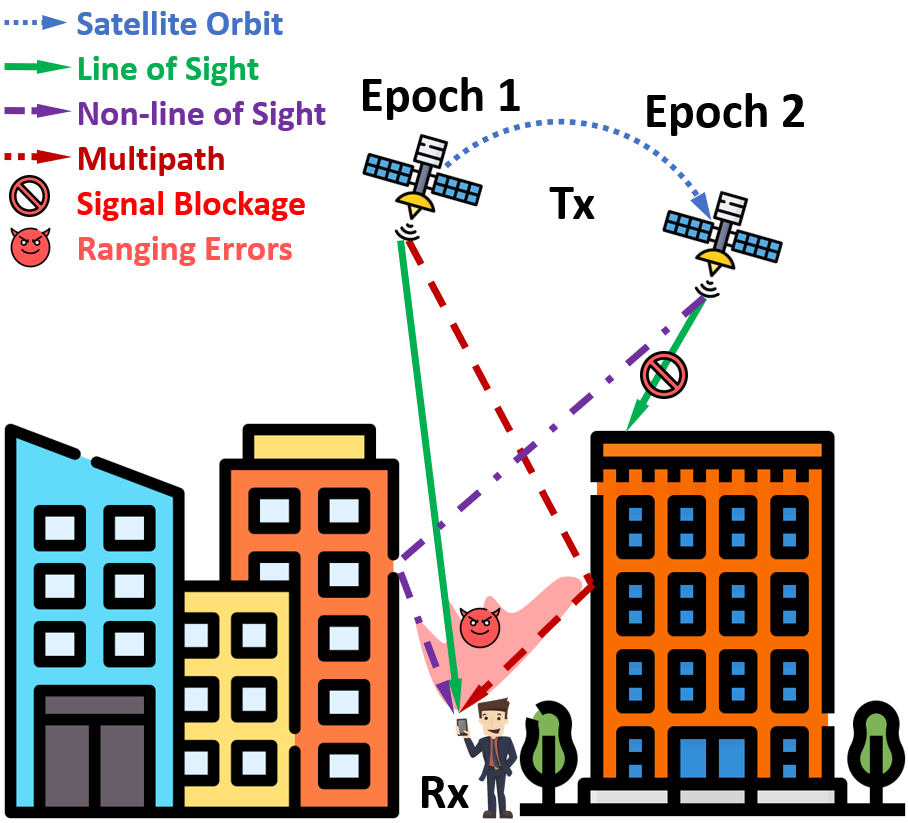}\label{fig:urban_nav}
}
  \hfil
  \subfloat[]{\includegraphics[width=0.48\linewidth]{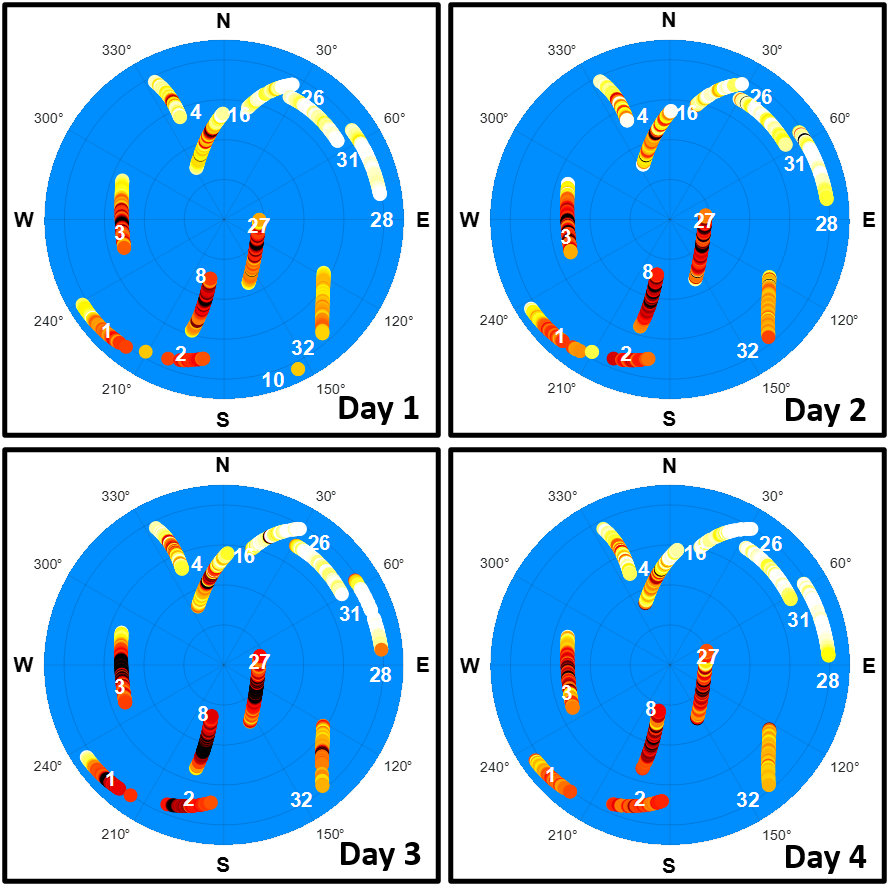}\label{fig:ranging_errors}}
  \caption{(a) Pervasive multipath/NLOS-induced ranging errors in urban environment. (b) Ranging errors of visible satellites observed at the same site and time across multiple days.}
  \Description{This picture shows that the multipath/NLOS effect is related to the relationship among users, satellites, and the environment.}
  \label{fig:neural_PrC}
\end{figure}

Global Navigation Satellite Systems (GNSS), a ranging-based wireless positioning system, equips mobile devices with global spatial awareness, serving as a cornerstone for numerous daily applications such as navigation, outdoor augmented reality (AR), city asset management, food delivery, and ride-hailing services. However, particularly for low-cost mobile devices, the complex urban environment poses significant challenges to positioning accuracy \cite{li2019characteristics,zhang2018quality}. For instance, as shown in \autoref{fig:urban_nav}, the prevalence of high-rise buildings often reflects or obstructs line-of-sight (LOS) GNSS signals, leading to multipath effects and non-line-of-sight (NLOS) propagation, both of which are longstanding issues that degrade GNSS ranging measurements and compromise positioning accuracy \cite{wang2015smartphone, 10.1145/3210240.3210343, ng2021urban,weng2023characterization}. Additionally, the suboptimal GNSS antennas and chipsets commonly found in commercial mobile devices exacerbate the problem \cite{zhang2018quality,li2019characteristics,liu2019quality,weng2023localization}. 


\begin{figure}[!t]
  \centering
  \includegraphics[width=\linewidth]{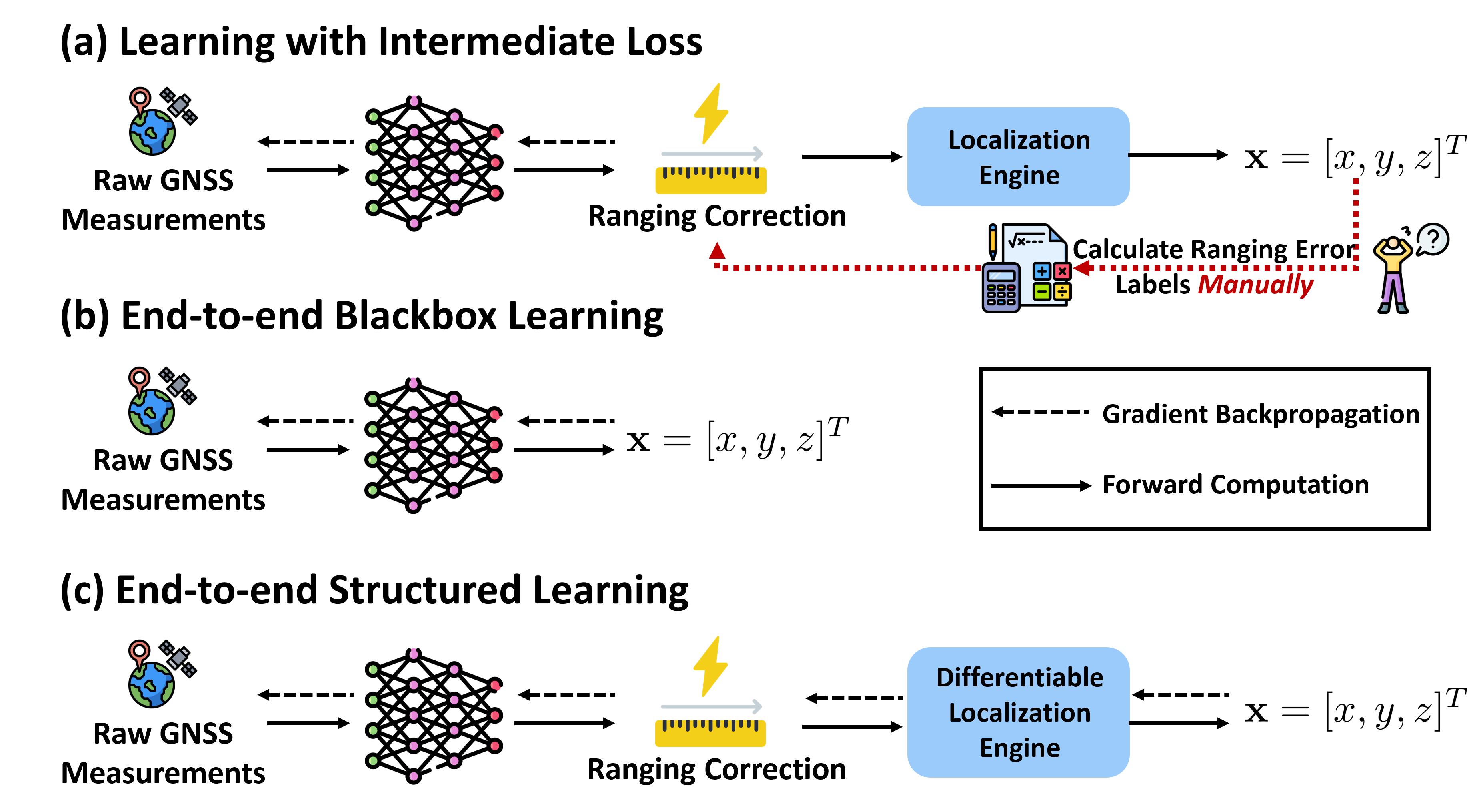}
  \caption{Categories of data-driven ranging error regression}
  \Description{This picture indicates three types of learning paradigms for the regression of pseudorange errors}
\label{fig:e2e_learning}
\end{figure}

The ranging errors caused by multipath and NLOS effects are strongly dependent on the complex relationship among satellites (Tx), receivers (Rx), and the surroundings \cite{sun2023resilient,10506762,sun20253d}. Although modeling errors in ranging measurements is challenging, they often exhibit a periodical pattern at a given site because GNSS satellites fly back approximately every 24 hours \cite{axelrad2005use,phan2013gps}. For example, we collected GNSS data using an Android phone at the same site and time across multiple days and visualized ranging errors in \autoref{fig:ranging_errors}. The color of each satellite’s trajectory indicates the value of ranging errors, estimated from ground-truth user locations after removing all known error sources, including satellite clock offsets, atmospheric delays, and relativistic effects \cite{10506762}. \autoref{fig:ranging_errors} shows that ranging errors at a specific site and time exhibit a similar pattern. 

The context-dependence and predictability of ranging errors have propelled the community away from traditional model-based approaches \cite{axelrad2005use,liu2018gnome,bai2020using} toward data-driven methods \cite{sun2020improving,sun2021using, 9490205,sun2023resilient,10506762,sun20253d}. However, obtaining accurate labels for ranging errors---critical for training neural models---is both difficult and impractical. As illustrated by \autoref{fig:e2e_learning}a, previous studies have used various sophisticated algorithms to estimate these errors based on ground truth locations provided by high-performance specialized positioning systems \cite{9490205,sun2023resilient,10506762,sun20253d}. With the derived ranging errors, an intermediate loss can be computed to train a neural network for ranging correction. Yet, manually calculating ranging errors is inherently imprecise due to the unobservable user clock offsets and stochastic noise \cite{10506762}. This raises the question: {\bfseries can we instead utilize these location labels directly, given that our ultimate goal is localization?} If so, we could bypass the need to compute intermediate losses using ranging error labels and instead train neural modules end-to-end with the final positioning task loss. 



\begin{figure}[!t]
\centering
{\includegraphics[width=\linewidth]{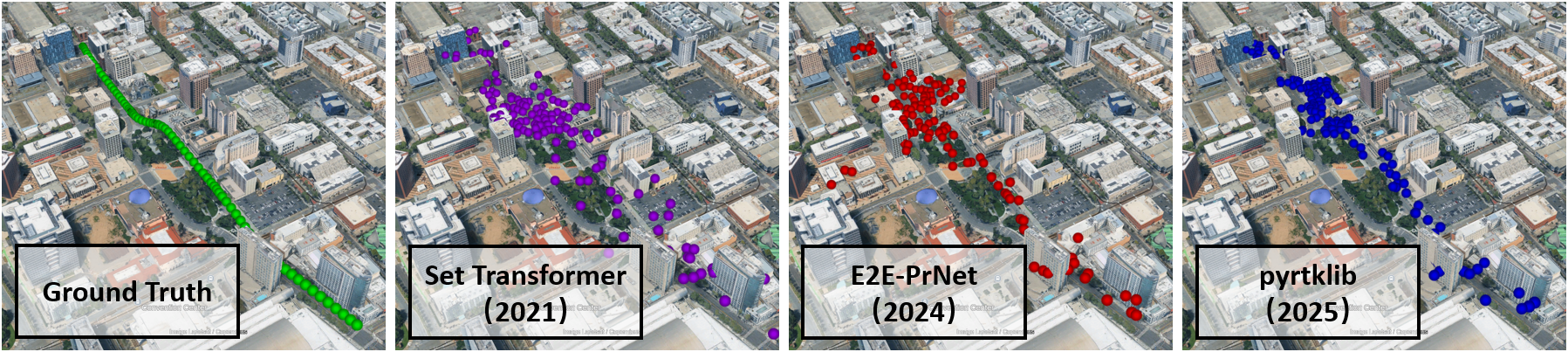}
}
\caption{Performance of SOTA end-to-end methods in cities 
}
\Description{The evolution of end-to-end approaches for GNSS positioning}
\label{fig:motivation}
\end{figure}

Motivated by this, researchers have proposed end-to-end neural models to map raw measurements to location corrections directly \cite{kanhere2022improving, mohanty2023learning,zhao2024improving}, as illustrated in \autoref{fig:e2e_learning}b. Nevertheless, replacing well-established physical or geometric priors with purely neural models turns them into black boxes with limited interpretability. In addition, their positioning accuracy is tightly constrained by the quality of the initial location estimates, to which the corrections are applied, making them susceptible to degraded performance in challenging areas \cite{mohanty2023learning,10506762}, as exemplified by the set transformer in \autoref{fig:motivation}.

Can we {\bfseries integrate} the two learning paradigms (shown in \autoref{fig:e2e_learning}a and 2b) to {\bfseries harness the powerful regression capabilities of neural networks while preserving our interpretable knowledge of GNSS localization?} The key ingredient to enable such an end-to-end learning framework should be a differentiable positioning engine, as illustrated in \autoref{fig:e2e_learning}c. This engine leverages outputs from an upstream neural model to compute locations while backpropagating gradients from final positioning tasks to refine the upstream neural model. This structured learning paradigm, widely adopted in fields such as robotics \cite{10313083, Yang-RSS-23, wang2025IL}, autonomous driving \cite{10154577}, and 3D vision \cite{10.1007/978-3-030-58452-8_24, chen2022epro}, has recently been extended to GNSS positioning \cite{xu2023differentiable,10706359,xu2024autow,hu2025pyrtklib}. 
Among the most relevant studies, E2E-PrNet and \verb|pyrtklib| train upstream neural networks to predict pseudorange errors through a differentiable Weighted Least Squares (WLS) localization engine \cite{10706359,hu2025pyrtklib}. While these methods improve positioning in rural \cite{10706359} or lightly urban areas \cite{hu2025pyrtklib}, they struggle in deep urban environments, as shown in \autoref{fig:motivation}. The main reason is that the WLS engine lacks robustness to the severe noise present in such settings, where multipath and NLOS propagation typically cause low carrier-to-noise density ratios ($C/N_0$) \cite{weng2023localization}. Furthermore, these methods rely solely on fully labeled 3D location data, limiting their practical deployment since obtaining accurate ground-truth locations is both challenging and costly \cite{fu2020android}.



These challenges raise a key question: {\bfseries can we build a neural ranging error correction framework trained end-to-end on the final location-related loss, while remaining robust to noise and capable of exploiting unlabeled data?} To address this, we propose a novel Neural Ranging Correction (NeRC) framework. NeRC is trained through differentiable moving horizon location estimation, offering robustness to noise, particularly in urban environments. Moreover, as blessings of end-to-end learning, we further enable training with 2D location data to relax the strict requirement for 3D location labels, and incorporate map-based supervision to unleash the power of unlabeled data. Our main contributions are:



\begin{itemize}[leftmargin=*]
\item  We design a sequence-to-sequence framework mapping a horizon of raw GNSS measurements to a trajectory of locations. The pipeline comprises an upstream MultiLayer Perceptron (MLP) for ranging error regression and a downstream differentiable localization engine based on Moving Horizon Estimation (MHE).



\item We justify the selection of MHE as the downstream location estimator and profile its forward and backward performance.

\item We propose training NeRC with only 2D location labels, and further enabling the use of unlabeled data by incorporating supervision from Euclidean Distance Field (EDF) cost maps.




\item We comprehensively evaluate NeRC on public smartphone benchmarks and our own collected data, demonstrating that NeRC reaches a new state-of-the-art (SOTA) level in GNSS positioning.


\item We deploy NeRC in an edge-based system to enable real-time ranging correction for mobile localization, and extensively profile the system performance, verifying its feasibility in the real world.  
\end{itemize}

To the best of our knowledge, this is the first work to learn pseudorange error representations through a differentiable localization engine capable of robust operation in both rural and urban environments. It is also the first to leverage unlabeled GNSS data using publicly available map information. Furthermore, we present a real-world, real-time demonstration of a learning-based mobile system for improving GNSS positioning.

\section{NeRC Design}
NeRC is a data-driven framework guided by well-established physical and geometric priors. It handles a horizon of raw measurements for neural ranging error regression and robust location estimation. Within it, the neural module is trained with positioning-related metrics. This end-to-end learning pipeline---from raw measurements to locations---comprises two core modules: 
\begin{itemize}[leftmargin=*]
\item{\bfseries Upstream Neural Network}: We utilize an MLP to map raw GNSS measurements to pseudorange errors, which serve as one of the inputs to the downstream localization module.

\item{\bfseries Downstream Differentiable Location Estimator}: We design a model-based positioning engine using MHE to handle neural ranging error predictions and track user trajectories robustly, even in noisy urban areas. Its differentiable formulation enables gradients from the location-related loss to propagate backward, allowing end-to-end training of the upstream network.
\end{itemize}

\begin{figure}[!t]
  \centering
  \includegraphics[width=\linewidth]{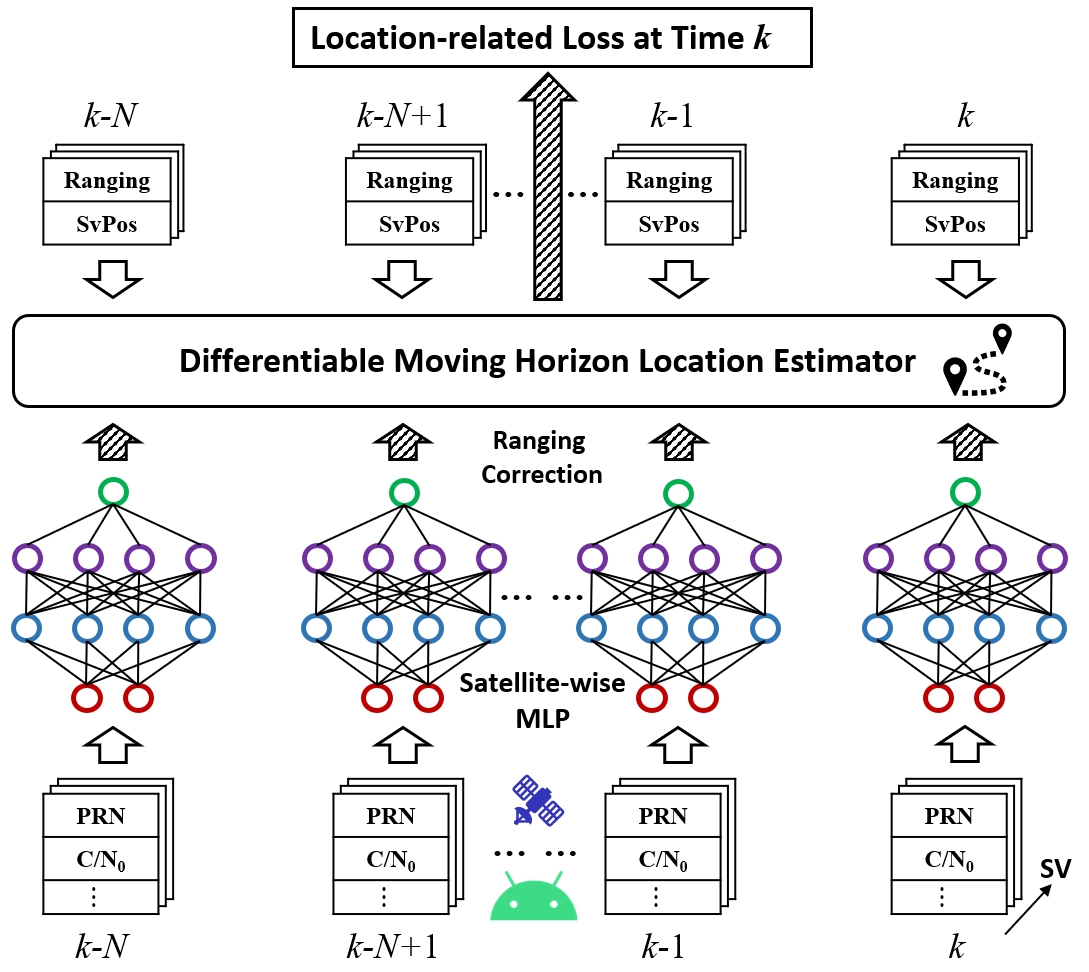}
  \caption{Principle diagram of NeRC.}
  \Description {NeRC is a sequence-to-sequence framework with raw GNSS measurements as inputs and localization results as outputs. A location-related loss can be computed as a result.}
  \label{fig:nerc_overview}
\end{figure}

\autoref{fig:nerc_overview} displays the principal structure of NeRC that processes a horizon of $N+1$ measurements at time $k$. $N$ is defined as the horizon size. At each time step, except for the feature dimension, the input data possesses another dimension representing different satellites. For example, as shown by the overlapped data portfolio at time $k$, the satellite dimension is denoted by SV, which has a size of 32 for the US's GPS constellation. 
In the following sections, the details of the diagram are explained.

\subsection{Upstream Network for Ranging Correction}
Considering the SOTA performance of PrNet---an MLP trained with the surrogate intermediate loss---in mobile device GNSS positioning \cite{10506762}, we use a similar MLP backbone as the upstream neural network in NeRC. The MLP is designed to learn the mapping from raw GNSS measurements to pseudorange errors. In addition to the input features mentioned in \cite{10506762}, we include another two features representing the quality of ranging measurements---pseudorange residuals $\varepsilon_r$ and their root-sum-squares values $E_r$ \cite{9490205}. In sum, the input features are denoted by a tensor as follows:
\begin{equation*}
\mathbf{I}=\left[C/N_0, \theta_e, {\rm PRN}, \hat{\mathbf{p}}_{\rm LLA}, \mathbf{g}_{\rm NED}, \hat{\mathbf{d}}_{\rm NED}, \varepsilon_r, E_r\right]^T
\end{equation*}
where $C/N_0$ is the carrier-to-noise density ratio, and $\theta_e$ denotes the elevation angles of visible satellites. PRN refers to the pseudorandom noise codes used as identifiers for GNSS satellites. $\hat{\mathbf{p}}_{\rm LLA}$ is the user position estimate from baseline methods like WLS, expressed in the longitude-latitude-altitude (LLA) frame, providing the neural network with awareness of the spatial distribution of ranging errors \cite{sun2023resilient,10506762,sun20253d}. $\mathbf{g}_{\rm NED}$ denotes the unit vectors from satellites to the user in the north–east–down (NED) frame, capturing the spatial satellite-user geometry. $\hat{\mathbf{d}}_{\rm NED}$ is the direction estimation using baseline positioning methods, which to some extent reflects the antenna orientation of mobile devices \cite{10506762}.

\autoref{fig:nerc_overview} illustrates that one input tensor sample has two dimensions corresponding to satellites and features. The MLP operates in a satellite-wise manner---analogous to the position-wise MLPs in the transformer architecture---processing information from all satellites in parallel. A visibility mask handles the time-varying set of visible satellites by zeroing the outputs of invisible ones, analogous to the transformer’s masked softmax. For clarity, \autoref{fig:nerc_overview} presents an unfolded view of the input tensors from time $k-N$ to time $k$. In reality, the NeRC pipeline uses a single MLP that processes the entire time window in parallel. Including the time dimension and the batch dimension, the resulting four-dimensional input tensor has the shape $B\times (N+1) \times {\rm SV} \times I$, where $B$ is the batch size, $\rm SV$ is the total number of satellites, and $I$ is the number of input features. Thus, the MLP can be represented by
\begin{equation}
\hat{\boldsymbol{\upepsilon}}_{k-N:k}=f(\mathbf{I}_{k-N:k};\boldsymbol{\Phi})\label{eqn:neurPrerr}
\end{equation}
where $\mathbf{I}_{k-N:k}$ denotes the 4D input tensor with a horizon size of $N$ at time $k$, $\hat{\boldsymbol{\upepsilon}}_{k-N:k}$ is the corresponding pseudorange error predictions, and $\boldsymbol{\Phi}$ represents all learnable parameters in the neural network.

\subsection{Downstream Location Estimator}

\subsubsection{Structured Learning Mechanism} Given the ranging correction $\hat{\boldsymbol{\upepsilon}}_{k-N:k}$ from the upstream MLP, as shown in \autoref{fig:nerc_overview}, combined with other necessary measurements $\boldsymbol{\Upsilon}_{k-N:k}$, including satellite positions, satellite velocities, pseudoranges, and pseudorange rates, the differentiable MHE estimates user trajectories $\hat{\mathbf{x}}_{k-N:k}$ from time $k-N$ to time $k$, which can be represented by
\begin{equation}
    \hat{\mathbf{x}}_{k-N:k}=g(\hat{\boldsymbol{\upepsilon}}_{k-N:k};\boldsymbol{\Upsilon}_{k-N:k}).\label{eqn:stateloss}
\end{equation}

Let $\mathcal{L}$ denote the location-related loss function. Then, we have the loss value $J_k$ at time $k$
\begin{equation}
    J_k = \mathcal{L}(\hat{\mathbf{x}}_{k-N:k}).\label{eqn:J_loss}
\end{equation}
Then, composed of \eqref{eqn:neurPrerr}, \eqref{eqn:stateloss}, and \eqref{eqn:J_loss}, the gradients of the loss with respect to the learnable parameters of the upstream neural network can be expressed using the chain rule:
\begin{equation}\label{eqn:gradients}
\frac{\partial J_k}{\partial \mathbf{\Phi}}=\frac{\partial J_k}{\partial \hat{\mathbf{x}}_{k-N:k}}\cdot\frac{\partial \hat{\mathbf{x}}_{k-N:k}}{\partial \hat{\boldsymbol{\upepsilon}}_{k-N:k}}\cdot\frac{\partial \hat{\boldsymbol{\upepsilon}}_{k-N,k}}{\partial \mathbf{\Phi}}.
\end{equation}
On the right side of \eqref{eqn:gradients}, the first item is normally easy to compute considering the explicit definition of the loss function $\mathcal{L}$, e.g., the mean squared error loss. The third derivative can be obtained using standard deep learning libraries, such as PyTorch or TensorFlow. The middle one, the derivative of user trajectories with respect to ranging corrections, is derived through differentiable MHE.

\subsubsection{Moving Horizon Estimation}
We aim to design a robust algorithm to estimate user locations in a GNSS-based system, a NonLinear Time-Variant (NLTV) system, from a bunch of noisy ranging measurements, which can be formulated as
\begin{eqnarray}
\mathbf{x}_{k+1}&=&f_k(\mathbf{x}_k)+\mathbf{w}_k \label{eqn:dynEqn}
\\
\mathbf{y}_k&=&h_k(\mathbf{x}_k)+\mathbf{v}_k\label{eqn:measEqn}
\end{eqnarray}
where \eqref{eqn:dynEqn} describes the user dynamics with the process noise term $\mathbf{w}_k$ representing the discrepancy between the modeled dynamics and the actual motion. The measurement equation \eqref{eqn:measEqn} relates ranging measurements $\mathbf{y}_k$ to the user state $\mathbf{x}_k$, with $\mathbf{v}_k$ representing measurement noise. The initial state of the system is assumed to be a priori, denoted as $\mathbf{x}_0\sim N(\bar{\mathbf{x}}_0, \mathbf{Q}_{-1})$.

MHE uses a fixed-size window of measurements that slides along the time axis to approximate the full-information state estimation---known as Batched Least Squares (BLS)---which leverages all measurements from the start up to the current time step \cite{muske1993receding,rao2003constrained}. MHE is formulated as
\begin{eqnarray}
\min_{\{\hat{\mathbf{w}}_{j|k}\}_{k-N-1}^{k-1}}\Phi_k&&=\underbrace{\hat{\mathbf{w}}_{k-N-1|k}^T\mathbf{P}_{k-N}^{-1}\hat{\mathbf{w}}_{k-N-1|k}}_{\text{Prior State Cost (Arrival Cost)}}+ \nonumber
\\
&&\underbrace{\sum_{j=k-N}^{k-1}\hat{\mathbf{w}}_{j|k}^T\mathbf{Q}_j^{-1}\hat{\mathbf{w}}_{j|k}}_{\text{State Transition Cost}}+\underbrace{\sum_{j=k-N}^{k}\hat{\mathbf{v}}_{j|k}^T\mathbf{R}_j^{-1}\hat{\mathbf{v}}_{j|k}}_{\text{Measurement Cost}}\nonumber
\end{eqnarray}
\begin{eqnarray}
s.t.:\hat{\mathbf{w}}_{k-N-1|k}&=&\hat{\mathbf{x}}_{k-N|k}-\hat{\mathbf{x}}_{k-N|k-N-1}\nonumber
\\
\hat{\mathbf{w}}_{j|k}&=&\hat{\mathbf{x}}_{j+1|k}-f_j(\hat{\mathbf{x}}_{j|k})\nonumber
\\
\hat{\mathbf{v}}_{j|k}&=&\hat{\mathbf{y}}_{j}-h_j(\hat{\mathbf{x}}_{j|k}) \label{eqn:mhe}
\end{eqnarray}
where $\hat{[\cdot]}_{j|k}$ denotes an estimate of an unknown at time $j$ given measurements up to time $k$. $\mathbf{Q}_j$ and $\mathbf{R}_j$ represent the covariance matrices of process and measurement noise, respectively. The inverse of them is used to weight the state transition and measurement costs. $\mathbf{P}_{k-N}$ is the covariance matrix of the state estimation $\hat{\mathbf{x}}_{k-N|k-N-1}$, representing prior knowledge of our confidence in the initial user state for the current horizon $[k-N, k]$. It also represents the ``inertia'' of the system---a larger $\mathbf{P}_{k-N}^{-1}$ increases the contribution of the prior state cost (arrival cost) in the overall objective, thereby drawing the state estimates in the current horizon more consistent with historical trajectories. Assuming the system statistics in \eqref{eqn:mhe} are Gaussian, MHE formulated by \eqref{eqn:mhe} is equivalent to the Maximum A Posteriori (MAP) estimation  \cite{jazwinski1970stochastic,weng2025recedinghorizonrecursivelocation}:
\begin{equation}\label{eqn:MAP}
    \max_{\{\hat{\mathbf{x}}_{j|k}\}_{k-N}^{k}} p(\mathbf{x}_{k-N},\cdots,\mathbf{x}_k|\mathbf{y}_{k-N},\cdots,\mathbf{y}_k).
\end{equation}

\begin{figure}[!t]
  \centering
  \includegraphics[width=\linewidth]{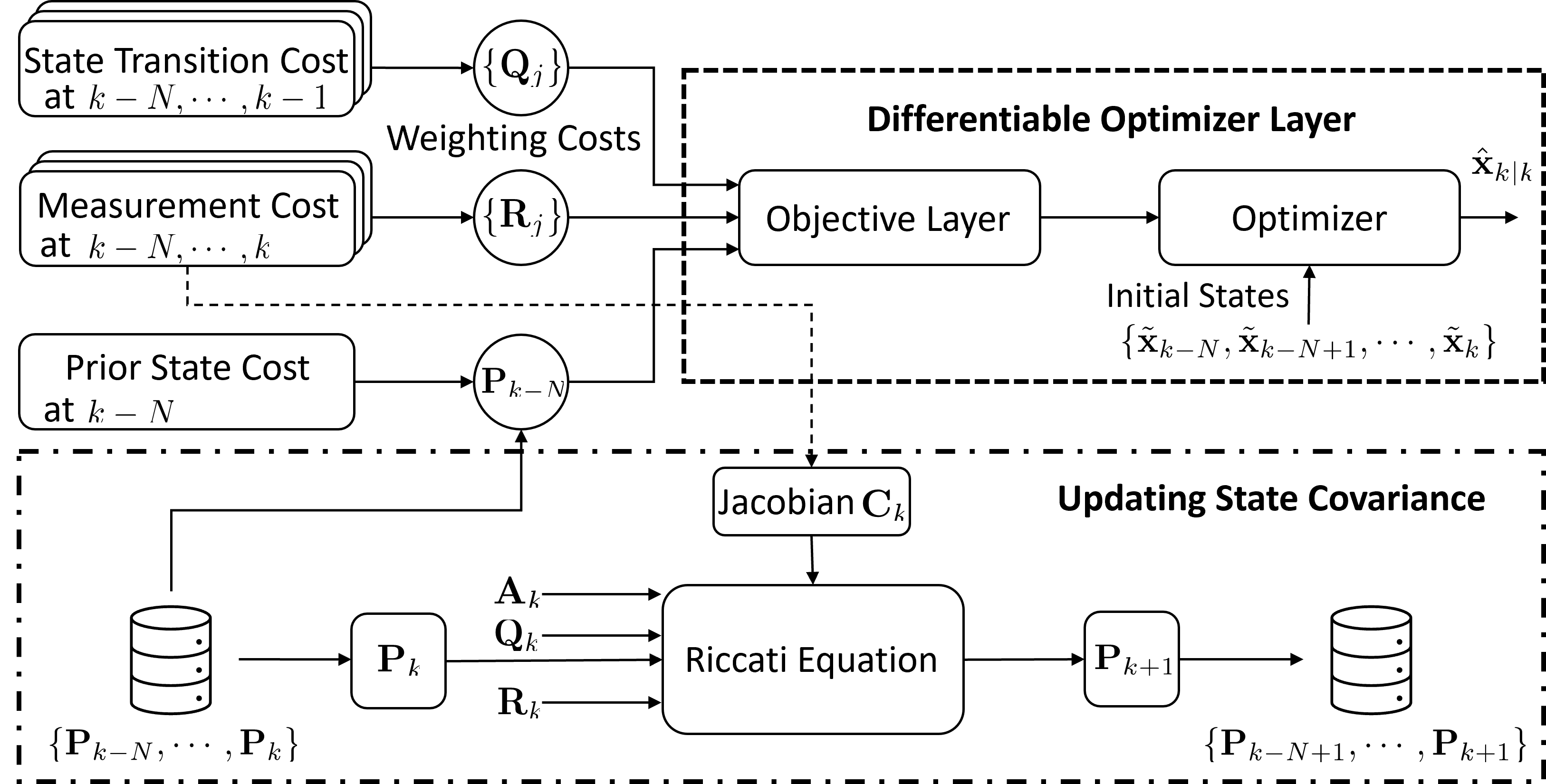}
  \caption{Principle diagram of differentiable MHE. It comprises three layers: the cost computation layer, the differentiable optimizer layer, and the state covariance update layer.}
  \Description{This figure describes the principle of moving horizon estimation.}
  \label{fig:dmhle_overview}
\end{figure}

\subsubsection{Differentiable Formulation of MHE}
We estimate the user location $\left[x_k, y_k, z_k\right]^T$, velocity $\left[v_{x_k}, v_{y_k}, v_{z_k}\right]^T$, receiver clock offset $\delta t_{u_k}$, and clock drift $\delta f_{u_k}$ simultaneously:
\begin{equation}\label{eqn:xk}
{\mathbf{x}}_k=\left[x_k, v_{x_k},y_k, v_{y_k},z_k, v_{z_k},\delta t_{u_k},\delta f_{u_k}\right]^T.    
\end{equation}
Considering the low dynamics of daily mobile devices, the system dynamics between consecutive time samples is typically simplified as uniform rectilinear motion. So, the nonlinear function $f_k(\cdot)$ in \eqref{eqn:dynEqn} becomes a linear coefficient matrix $\mathbf{A}_k$:
\begin{equation}\label{eqn:Ak}
f_k(\cdot)=\mathbf{A}_k=diag\{\mathbf{A}_{0_k},\mathbf{A}_{0_k},\mathbf{A}_{0_k},\mathbf{A}_{0_k}\},\quad\mathbf{A}_{0_k}=\begin{bmatrix}1&T_{k}
\\
0&1
\end{bmatrix}
\end{equation}
where $T_{k}$ is the sampling interval between time $k$ and $k+1$. Two types of ranging measurements are embodied by $\mathbf{y}_k$, i.e., pseudoranges $\rho_k^{(n)}$ and pseudorange rates $\dot{\rho}_k^{(n)}$, where $[\cdot]_k^{(n)}$ denotes the measurement of the satellite $n$ at time $k$. Pseudorange rates are normally accurate enough for velocity estimation, so we only regress pseudorange errors $\epsilon_k^{(n)}$ for better location estimation. {\bfseries The satellite clock offsets, relativistic effects, and atmospheric delays were removed from pseudorange measurements during data preprocessing \cite{kaplan2017understanding}}. Therefore, $\mathbf{y}_k$ is written as:
\begin{equation}\label{eqn:yk}
\left[\rho_k^{(1)}-\hat{\epsilon}_k^{(1)},\dot{\rho}_k^{(1)},\cdots,\rho_k^{(M)}-\hat{\epsilon}_k^{(M)},\dot{\rho}_k^{(M)}\right]^T
\end{equation}
where $M$ is the total number of visible satellites at time $k$. The relationship between the user state and ranging measurements, the nonlinear function $h_k(\cdot)$ in \eqref{eqn:measEqn}, can be modeled as \cite{morton2021position}:
\begin{eqnarray}
\rho_k^{(n)}&=&||\mathbf{p}_k-\mathbf{p}_k^{(n)}||+\delta t_{u_k}\label{eqn:range}
\\
\dot{\rho}_k^{(n)}&=&(\mathbf{v}_k-\mathbf{v}_k^{(n)})\cdot\mathbf{g}_k^{(n)}+\delta f_{u_k}\label{eqn:rangerate}
\end{eqnarray}
where $\mathbf{p}_k^{(n)}$ and $\mathbf{v}_k^{(n)}$ represents the location and velocity of satellite $n$ at time $k$. $\mathbf{g}_k^{(n)}$ is the unit geometry vector from satellite $n$ to the user at time $k$, which is computed as:
\begin{equation*}
\mathbf{g}_k^{(n)}=\left[{\tilde{x}_k-x_k^{(n)}},{\tilde{y}_k-y_k^{(n)}},{\tilde{z}_k-z_k^{(n)}}\right]^T/{\tilde{r}_k^{(n)}}
\end{equation*}
where $\tilde{\mathbf{p}}_k = \left[\tilde{x}_k,\tilde{y}_k,\tilde{z}_k\right]^T$ is the approximate user location guess for system initialization. $\tilde{r}_k^{(n)}$ denotes the geometry distance between the satellite $n$ and the location $\tilde{\mathbf{p}}_k$.




We adaptively compute the covariance matrix $\{\mathbf{Q}_j\}_{k-N}^{k-1}$ of process noise using a two-state model with two integrals that transfer disturbance (or accelerations) to state displacements \cite{spilker1996global}. For the measurement noise covariance matrix $\{\mathbf{R}_j\}_{k-N}^{k}$, we construct it as a diagonal matrix by placing the 1-sigma uncertainties of ranging measurements provided by mobile devices on its main diagonal. To approximate the full-information estimator that uses measurements since time zero, the covariance matrix $\mathbf{P}_{k-N}$ is computed recursively using the linear discrete filtering Riccati equation \cite{rao2003constrained}:
\begin{equation}\label{eqn:riccati}
\mathbf{P}_{k+1}=\mathbf{Q}_{k}+\mathbf{A}_{k}[\mathbf{P}_{k}-\mathbf{P}_{k}\mathbf{C}_{k}^T(\mathbf{C}_{k}\mathbf{P}_{k}\mathbf{C}_{k}^T+\mathbf{R}_{k})^{-1}\mathbf{C}_{k}\mathbf{P}_{k}]\mathbf{A}_{k}^T
\end{equation}
where $\mathbf{P}_{0}$ is treated as prior knowledge and is initialized as a diagonal matrix with all diagonal elements tuned empirically. $\mathbf{C}_k$ is the Jacobian matrix of the nonlinear function $h_k(\mathbf{x}_k)$:
\begin{equation}
\mathbf{C}_k=\begin{bmatrix}\tilde{a}_{x_k}^{(1)}&0&\tilde{a}_{y_k}^{(1)}&0&\tilde{a}_{z_k}^{(1)}&0&1&0
	\\
	0&\tilde{a}_{x_k}^{(1)}&0&\tilde{a}_{y_k}^{(1)}&0&\tilde{a}_{z_k}^{(1)}&0&1
	\\
	\vdots&\vdots&\vdots&\vdots&\vdots&\vdots&\vdots&\vdots
	\\
	\tilde{a}_{x_k}^{(M)}&0&\tilde{a}_{y_k}^{(M)}&0&\tilde{a}_{z_k}^{(M)}&0&1&0
	\\
	0&\tilde{a}_{x_k}^{(M)}&0&\tilde{a}_{y_k}^{(M)}&0&\tilde{a}_{z_k}^{(M)}&0&1
\end{bmatrix}	
\end{equation}
where $\tilde{a}_{x_k}^{(n)}$, $\tilde{a}_{y_k}^{(n)}$, and $\tilde{a}_{z_k}^{(n)}$ are the three elements of $\mathbf{g}_k^{(n)}$ in turn. The prior state $\hat{\mathbf{x}}_{k-N|k-N-1}$ is considered known for the current horizon at time $k$ and computed using \eqref{eqn:dynEqn} given the previous state estimate $\hat{\mathbf{x}}_{k-N-1|k-N-1}$. Now, we have all the parts required for the moving horizon location estimation.

\autoref{fig:dmhle_overview} illustrates the principle of the differentiable MHE. By substituting \eqref{eqn:xk}$\sim$ \eqref{eqn:rangerate} into \eqref{eqn:mhe}, we obtain the state transition cost, measurement cost, and prior state cost. The objective layer aggregates the three types of costs weighted by their corresponding covariance matrices, which is then solved by a general Differentiable Nonlinear Least Squares (DNLS) solver. In our implementation, we employ the Theseus library built for Pytorch \cite{pineda2022theseus}. The optimizer is initialized with an approximate trajectory guess, denoted by $\{\tilde{\mathbf{x}}_{k-N},\cdots,\tilde{\mathbf{x}}_{k}\}$, for each horizon. We leverage the historical state estimates $\{\hat{\mathbf{x}}_{k-N|k-N-1},\cdots,\hat{\mathbf{x}}_{k|k-1}\}$ for initialization \cite{weng2025recedinghorizonrecursivelocation}. The lower half of this diagram depicts the updating process of the prior state covariance matrix using \eqref{eqn:riccati}, where only the latest covariance matrix $\mathbf{P}_{k}$ gets updated at time $k$. The updated covariance matrices $\{\mathbf{P}_{k-N+1},\cdots,\mathbf{P}_{k+1}\}$ serves the next $N+1$ horizons.


In this section, instead of intuitively using a sliding window mechanism, we design a localization engine based on the state space model and moving horizon estimation. In the following section, we profile its forward and backward performance.


\begin{figure*}[!t]
\centering
\subfloat[]{\includegraphics[width=0.24\textwidth]{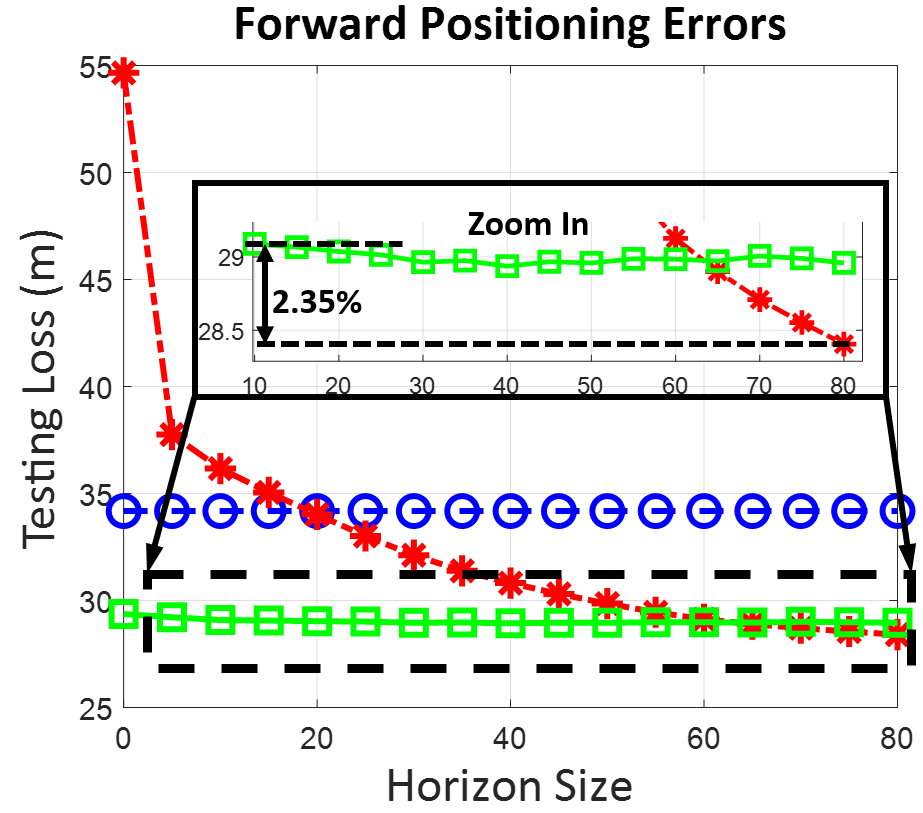}\label{fig:rmse3D}
}
  \hfil
\subfloat[]{\includegraphics[width=0.24\textwidth]{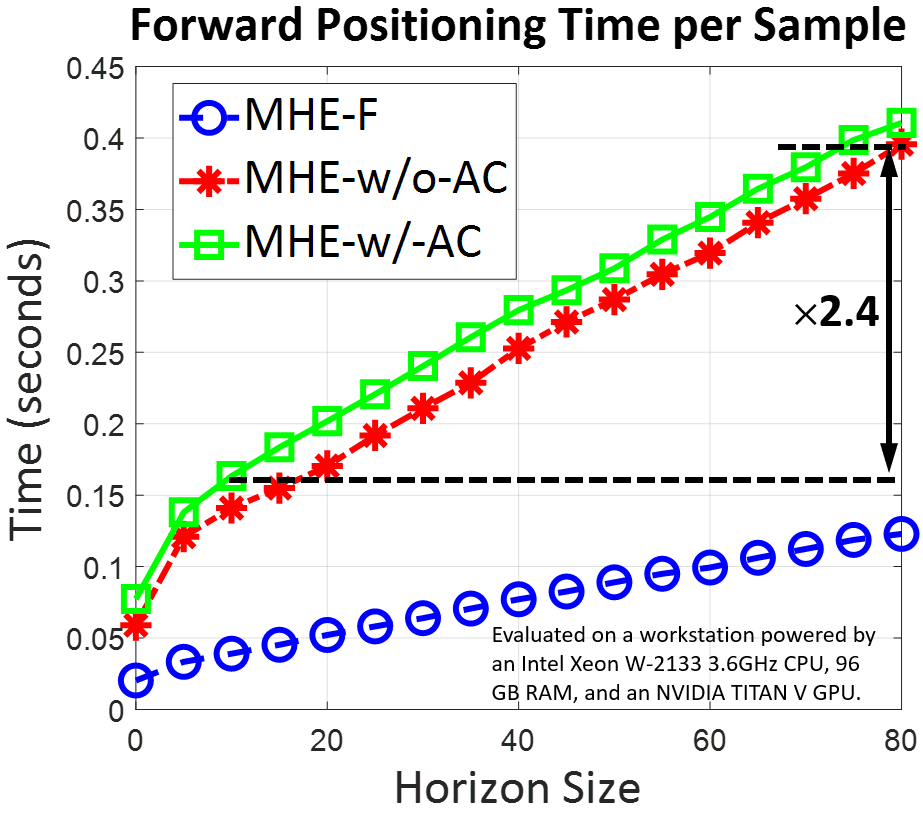}\label{fig: forwardTime}}
\hfil
\subfloat[]{\includegraphics[width=0.24\textwidth]{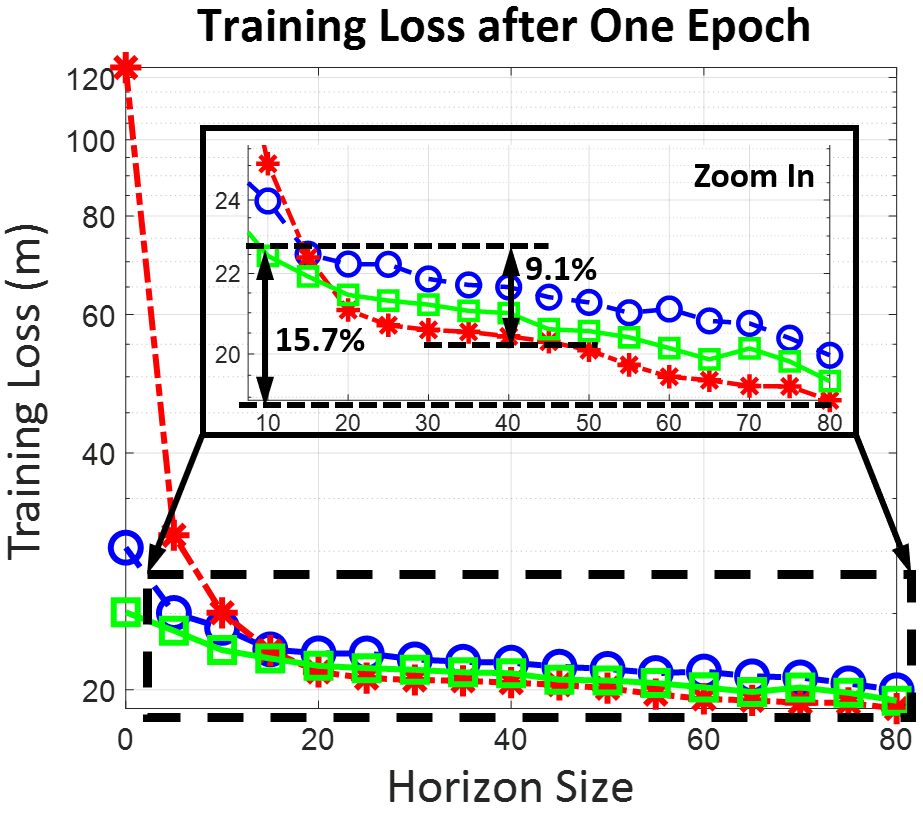}\label{fig:trainingLossN}
}
  \hfil
\subfloat[]{\includegraphics[width=0.24\textwidth]{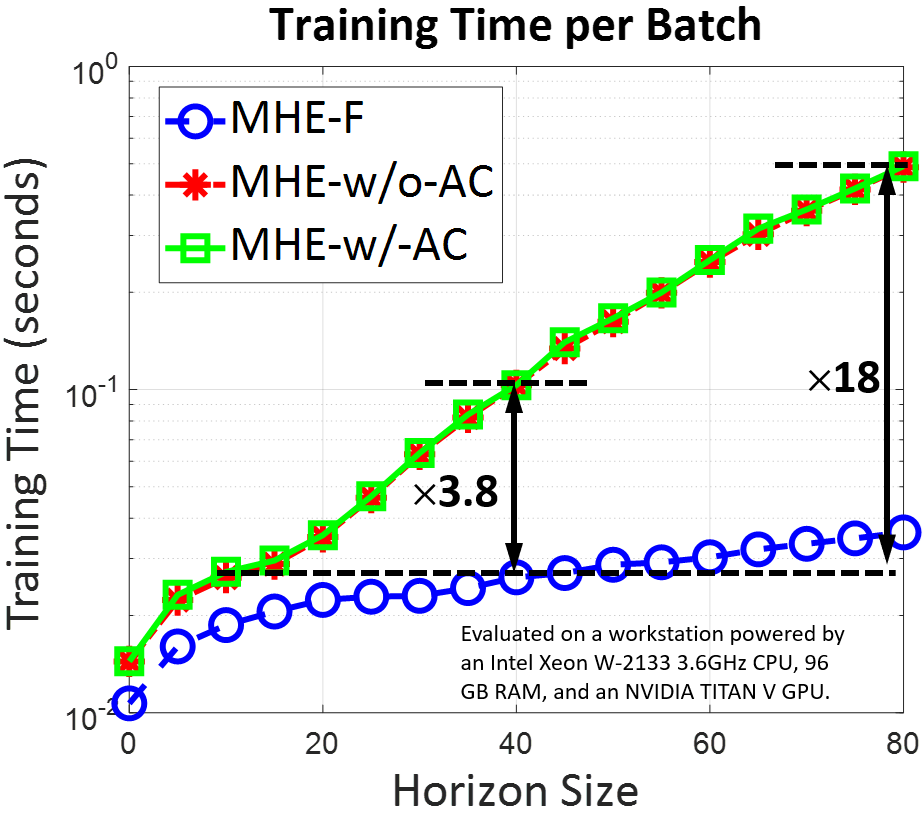}\label{fig: backwardTime}}
\caption{Forward and backward profiling of the filtering-based MHE (MHE-F), the MHE without arrival cost (MHE-w/o-AC), and the MHE with arrival cost (MHE-w/-AC) under various horizon sizes.} 
\Description{These figures profile the forward inference and backward training performance of MHE.}
\label{fig:profiling}
\end{figure*}

\section{Unification and Profiling}
\subsection{Unification}
{\bfseries Why do we choose MHE as the positioning engine rather than the classic Extended Kalman Filter (EKF) or the popular Factor Graph Optimization (FGO)?} The underlying reason is that the MHE framework can unify the other two algorithms. Their equivalence can be established as follows:
\begin{itemize}[leftmargin=*]
\item {\bfseries EKF}: Under certain prerequisites about the initial trajectories and disturbance statistics of an NLTV system, a recursive expression for MHE can be analytically derived, formulating a filtering-based MHE and establishing the equivalence between them, regardless of the horizon size \cite{weng2025recedinghorizonrecursivelocation}.

\item {\bfseries FGO}: If the statistics of an NLTV system are Gaussian, FGO will be equivalent to MHE \cite{jazwinski1970stochastic}. However, in GNSS localization, FGO is typically implemented in the absence of the prior state factor \cite{wen2021factor,zhong2022real,xu2023differentiable}. In the following analysis, FGO refers to an MHE without the arrival cost.
\end{itemize}



Therefore, we provide a comparative study of {\bfseries the filtering-based MHE (MHE-F)}, {\bfseries the optimization-based MHE without the arrival cost (MHE-w/o-AC)}, and {\bfseries the optimization-based MHE with the arrival cost (MHE-w/-AC)} in terms of forward positioning and backward training in the following sections. For a fair comparison, they are implemented with the {\bfseries same} system model, disturbance statistics, and initial conditions. 
A public dataset collected by Pixel 4 in deep urban regions is used for evaluation. 

\subsection{Forward profiling}





First, we profile them as standalone positioning engines, without any neural modules involved. The forward testing loss (positioning errors) and computational time are shown in \autoref{fig:rmse3D} and 6b.


\autoref{fig:rmse3D} indicates that MHE-w/-AC maintains a lead over other algorithms as the horizon size is small. By contrast, MHE-w/o-AC leverages measurements within the current horizon and disregards prior information, which significantly degrades its performance in short horizons. Increasing the horizon size can narrow down the performance gap between MHE-w/-AC and MHE-w/o-AC. Once the horizon extends beyond a certain threshold---approximately 65 in this example---MHE-w/o-AC begins to outperform MHE-w/-AC. At that point, the window is sufficiently long to retain informative measurements for current-state estimation, while older data outside the window degrades performance \cite{ling1996state,ling1999receding,weng2025recedinghorizonrecursivelocation}. Continuously increasing the horizon size does not always lead to improved performance. A typical performance curve of MHE-w/o-AC initially decreases to a minimum, then gradually converges back toward that of MHE-w/-AC \cite{rao2002constrained, weng2025recedinghorizonrecursivelocation}. In this dataset, we observe that the turning point occurs at a horizon length of approximately 450. The filtering-based MHE-F exhibits the same performance regardless of the chosen horizon size \cite{muske1993receding,weng2025recedinghorizonrecursivelocation}. 

Regarding the computation time of each method, \autoref{fig: forwardTime} shows that the forward time increases as the horizon size enlarges. Among the methods, MHE-F demonstrates the fastest performance, benefiting from its recursive filtering structure. In contrast, the optimization-based approaches are naturally slower, as they solve a nonlinear least squares problem at each time step. Notably, MHE-w/-AC incurs slightly more computation than MHE-w/o-AC for the same horizon size, due to the additional optimization for the arrival cost. However, MHE-w/-AC can achieve superior positioning accuracy with a smaller horizon, thereby taking less computational time to reach similar performance levels.



{\bfseries Summary:} MHE can offer advanced positioning capabilities at a competitive cost with appropriate configurations of its horizon size and arrival cost. Increasing the horizon size can enhance the accuracy of MHE-w/o-AC, but the marginal gains over MHE-w/-AC with a smaller horizon come with a substantial rise in computational time---a 2.35\% improvement in accuracy comes at the expense of approximately 2.4 times the computation time, as shown in \autoref{fig:rmse3D} and \autoref{fig: forwardTime}. This trade-off is critical to consider when deploying MHE on resource-constrained platforms such as mobile devices. Additional improvement in MHE is outside the scope of this paper, but can be expected by approximating the arrival cost more accurately \cite{rao2000nonlinear,tenny2002efficient, rao2003constrained, fiedler2020probabilistic} and incorporating constraints on the system \cite{rao2001constrained, haseltine2005critical, wen2021factor, suzuki2023precise, xia2024integrity}.

\subsection{Backward profiling} 
This section examines the training convergence of NeRC, an upstream MLP equipped with MHE. \autoref{fig:trainingLossN} gives the training loss (3D location loss) after one training epoch for various horizon sizes. 

Increasing the horizon size improves training convergence across all estimators. While MHE-w/o-AC exhibits poor convergence at small horizons, it surpasses the other two methods in convergence speed once the horizon exceeds 20, as illustrated in \autoref{fig:trainingLossN}. It is the arrival cost item in \eqref{eqn:mhe}---the system's `inertia'---that decelerates the backward convergence of MHE-F and MHE-w/-AC. Interestingly, it is the same item that enables MHE to achieve satisfactory forward positioning accuracy even with small horizon sizes. 


\autoref{fig: backwardTime} shows the backward training time per batch of the three methods. Just aligned with the forward time, MHE-F exhibits the lowest backward computation time due to its filtering structure.

{\bfseries Summary:}
The backward training convergence can be improved by expanding the receding horizon, which is equivalent to reducing the contribution of the arrival cost in the overall optimization objective \eqref{eqn:mhe}. However, increasing the horizon size extends the training time. For instance, achieving a 9.1\% gain in training loss convergence requires approximately 3.8 times more training time, while a 15.7\% improvement demands nearly an 18-fold increase in training time. As the horizon grows, the marginal gain in convergence shrinks while the cost in training time grows steeply, which is a trade-off that requires a balance. 


\section{Training NeRC}
{\bfseries We train a separate NeRC for each area or route,} just like other data-driven regression paradigms that tightly depend on the specific scenes in which they are applied \cite{10.1007/978-3-030-58452-8_24,zhao2023nerf2,zhao2024crowdsourced}. Additionally, {\bfseries separate NeRC models should be trained for specific time intervals of a day}, as GNSS satellites (Tx) orbit the Earth periodically. Since pseudorange errors are also closely related to device variations \cite{10506762}, NeRC needs to be retrained for different mobile phones. We specify how to train NeRC here. In addition to supervised learning using data labeled with true locations, we propose an alternative training paradigm to unleash the power of unlabeled data.





\subsection{Supervised Learning}
It is standard and direct to use ground truth locations to train an end-to-end learning pipeline for localization \cite{kanhere2022improving,10706359}. The supervised training loss is shown as follows.
\begin{equation}\label{eqn: 3DLoss} 
\mathcal{L}_{3D} = ||\hat{\mathbf{p}}_k-\mathbf{p}_k||^2
\end{equation}
where the location label $\mathbf{p}_k$ can be collected using high-performance geodetic integrated positioning systems like NovAtel's SPAN system \cite{fu2020android, hsu2023hong}. The traditional data-driven ranging correction methods, which use intermediate training loss, also require labels of user clock offsets \cite{9490205,sun2023resilient,10506762}. However, tracking user clock offsets is challenging \cite{xu2019intelligent, 10506762}. By contrast, {\bfseries as a benefit of end-to-end learning for localization}, NeRC can be trained without them. It is also worth noting that supervising user velocities and clock drifts is unnecessary because the corresponding ranging rate measurements are accurate enough, requiring no further corrections.





Similarly, we can further eliminate the vertical location labels (altitudes) from the training process for two main reasons:
\begin{itemize}[leftmargin=*]
    \item It is also challenging to label the altitude, even with specialized high-performance positioning systems \cite{kaplan2017understanding}.
    \item We are more concerned about the horizontal positioning performance of mobile devices.
\end{itemize}
In GSDC 2022, Google removed the vertical labels and pushed researchers to focus on horizontal localization. As a blessing of end-to-end learning, NeRC allows the training with only 2D horizontal labels, thereby avoiding the use of inaccurate labels for altitudes. Let $\phi_k$ and $\lambda_k$ respectively denote the ground truth latitude and longitude of a user at time $k$. Then, we get the 2D training loss:
\begin{equation}\label{eqn: 2DLoss} 
\mathcal{L}_{2D}=||[\hat{\phi}_k,\hat{\lambda}_k]^T-[\phi_k,\lambda_k]^T||^2
\end{equation}
where
\begin{equation*}
[\hat{\phi}_k,\hat{\lambda}_k]^T=f_{\text{ECEF2LLA}}(\hat{\mathbf{p}}_k)
\end{equation*} 
where $f_{\text{ECEF2LLA}}$ represents the conversion of user locations from the ECEF coordinate system to the LLA frame, which is differentiable and can be integrated into the learning pipeline \cite{kaplan2017understanding}.

\begin{figure}[!t]
\centering
\subfloat[]{\includegraphics[width=0.225\textwidth]{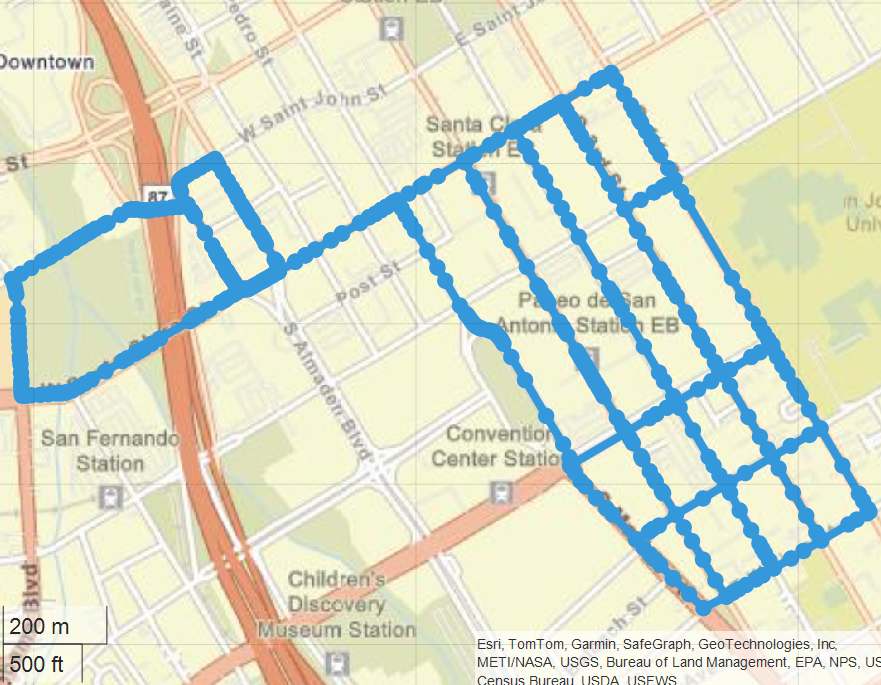}\label{fig:Pgp_raw_routes_myMaps}
}
  \hfil
\subfloat[]{\includegraphics[width=0.24\textwidth]{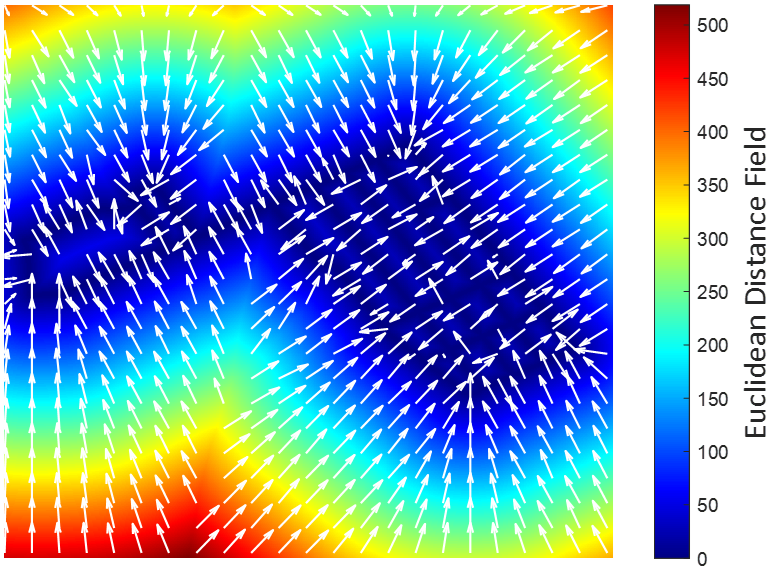}\label{fig: Pgp_esdf}}
\caption{(a) Routes extracted from Google Maps and (b) its Euclidean distance field cost map. The white arrows indicate the gradient descending directions of the cost.
}
\Description{These figures illustrate the publicly available route information from Google My Maps and the EDF cost map derived from the map information.}
\label{fig:esdf}
\end{figure}

\subsection{Unleashing the Power of Unlabeled Data}
Innumerable unlabeled GNSS data are generated from mobile devices every day, but have never been fully used. {\bfseries As the blessings of end-to-end learning}, we propose a brand new learning paradigm using Euclidean Distance Field (EDF) cost maps to unleash the power of unlabeled GNSS data. The core idea behind this innovation is to construct a differentiable map supervision by comparing the predicted trajectories with the passable routes provided by publicly available geodetic maps, like Google Maps or OpenStreetMap. This novel learning paradigm is inspired by robot motion planning \cite{ratliff2009chomp, han2019fiesta, Yang-RSS-23}, robust flight control (actual–reference trajectory comparison) \cite{10313083}, and map matching \cite{paglieroni1992distance,liu2022deepgps}. 

We assume that a user travels along the passable routes in an area. As shown in \autoref{fig:Pgp_raw_routes_myMaps}, we can extract passable routes from Google Maps and construct a corresponding spatial cost function. The design principle is to ensure a lower cost value as the user's trajectory becomes more aligned with the reference route. To this end, we propose to employ the Euclidean distance from each predicted location to its nearest reference route as the cost to supervise the learning process. This can be done through the Euclidean distance transform (EDT) of the map as illustrated by \autoref{fig: Pgp_esdf}, which is referred to as the Euclidean distance field (EDF) cost map \cite{han2019fiesta, Yang-RSS-23}. For each spatial point $\mathbf{p}$ on the map, EDT produces a potential value $\mathcal{D}(\mathbf{p})$ representing the distance between $\mathbf{p}$ and its nearest point on the reference route $\mathcal{G}$ \cite{felzenszwalb2012distance}: 
\begin{equation*}
\mathcal{D}(\mathbf{p})=\min_{\mathbf{q}\in\mathcal{G}}\left(||\mathbf{p}-\mathbf{q}||_2\right)
\end{equation*}
where $\mathbf{q}$ denotes a point on the reference route $\mathcal{G}$. For the user trajectory in the horizon at time $k$, the EDF training loss is:
\begin{equation}\label{eqn:edf_Loss}
    \mathcal{L}_{EDF}=\frac{1}{N+1}\sum_{i=k-N}^{k}\mathcal{D}\left(\hat{\mathbf{p}}_i\right).
\end{equation}
The gradient descending directions of the cost map are illustrated by white arrows in \autoref{fig: Pgp_esdf} and always push location predictions toward reference trajectories, which is a dual analogy for collision avoidance in robot motion planning \cite{ratliff2009chomp, han2019fiesta, Yang-RSS-23}. Even when data collection routes slightly deviate from reference trajectories, considerable improvement can still be achieved, since GNSS positioning errors in urban areas often span hundreds of meters, exceeding street widths. Note also that EDF cost maps serve here only as loss functions for training with unlabeled data, not as optimization constraints. And only one EDF cost map is needed to train NeRC for a given region across different times of day. The inference process does not require any map information.


\begin{figure}[!t]
  \centering
  \includegraphics[width=\linewidth]{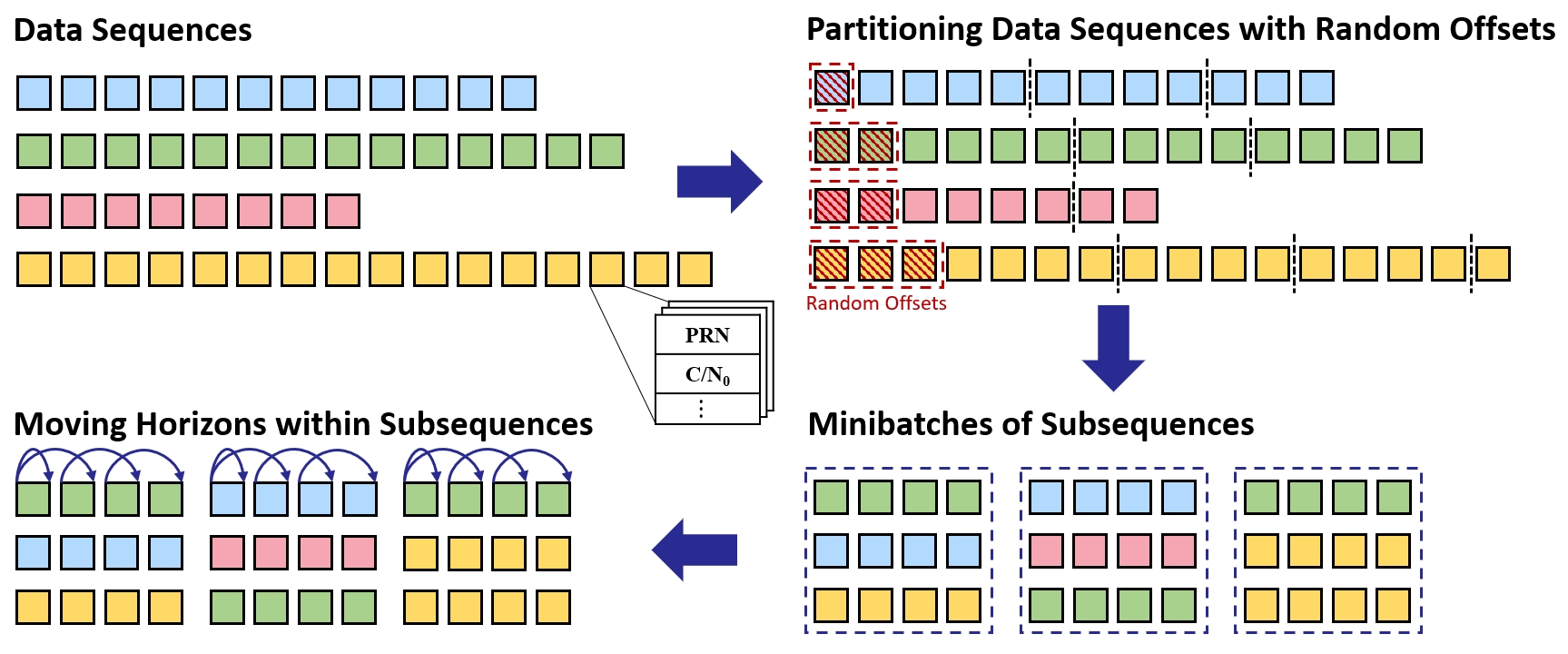}
  \caption{Sequential data loader for training NeRC} 
  \Description{This picture shows the sequential data loader for training NeRC.}
  \label{fig:dataloader}
\end{figure}

\section{NeRC Implementations}
NeRC is a sequence-to-sequence data-driven framework comprising an upstream MLP and a differentiable downstream MHE. We leverage PyTorch and Theseus to implement them. The specific configurations are listed below.

\subsection{Sequential Data Loader}
Drawing inspiration from the techniques for loading long sequences in natural language processing \cite{zhang2023dive}, we design a sequential data loader tailored for handling raw GNSS measurements, as illustrated in \autoref{fig:dataloader}. Long sequences from data files are first truncated using random offsets and subsequently divided into equal-length subsequences. These subsequences serve as fundamental units, which are then randomly assembled into mini-batches. Finally, a sliding window moves along the batched subsequences to perform MHE.


\subsection{MLP Configuration} 
We adopt the parameter settings of the SOTA PrNet \cite{10506762} to build the upstream neural network---a 40-layer MLP with 20 hidden neurons per layer, using ReLU as the activation function. The MLP is initialized with values drawn from a Kaiming normal distribution \cite{he2015delving}. For training the upstream network, we use the Adam optimizer with a learning rate that starts at 0.01 and decays over time. Training NeRC is typically efficient, converging within minutes to a few hours, depending on the scale of the training data.

\subsection{MHE Configuration}
The differentiable MHE is empowered by Theseus, a DNLS solver built for PyTorch \cite{pineda2022theseus}. The Gauss-Newton optimizer is selected as the optimization kernel, with the maximum number of iterations and step size empirically tuned for optimal performance. Specifically, we set the step size to 0.5 and limit the number of iterations to 10. To balance the trade-off between training performance and efficiency, as indicated by \autoref{fig:profiling}, we eliminate the arrival cost and set the horizon size to 15. During inference, we employ the MHE with the arrival cost and set the horizon size to 5. 





\subsection{Building Euclidean Distance Cost Maps} \label{App: esdf}

\begin{figure}[!t]
  \centering
  \includegraphics[width=\linewidth]{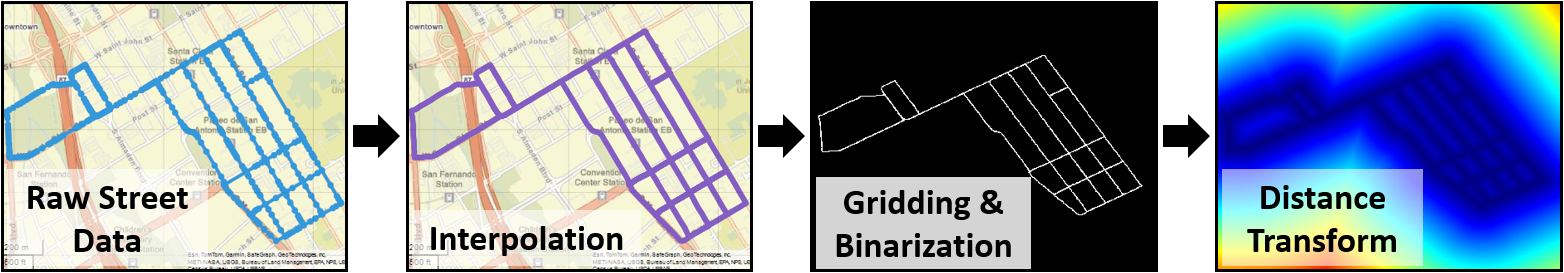}
  \caption{The pipeline of Euclidean distance field cost map computation. Raw street data is from Google My Maps.}
  \Description{The flowchart to construct EDF cost maps}
  \label{fig: esdf_workflow}
\end{figure}

The workflow of constructing an EDF cost map is illustrated by \autoref{fig: esdf_workflow}. We start with collecting raw reference street data using Google My Maps, where walking, biking, and driving routes are freely available. Such data can be exported as a file of waypoints in the Keyhole Markup Language (KML) format. However, the raw street waypoints are too sparse to provide efficient supervision. Thus, we leverage the cubic spline to interpolate these raw route waypoints to obtain a dense and smooth reference route. Then, we grid the interpolated map on the longitude-latitude lattice with a tunable resolution. The grid cells occupied by the passable route are set to 1, while the blocked cells are set to 0. In this way, the gridded map turns into a binary image for efficient EDT. Many mature solutions to the EDT of binary images are available \cite{paglieroni1992distance,breu1995linear, maurer2003linear,felzenszwalb2012distance}. We use the \verb|bwdist| method provided by MATLAB to compute the EDF of the gridded map. To make the EDF cost map more differentiable, we apply to the map matrix a Gaussian lowpass filter of size 5 with a standard deviation of 1 \cite{Yang-RSS-23}. Then, we compute the loss function \eqref{eqn:edf_Loss} using PyTorch's \verb|grid_sample| method.

\section{Evaluation}
This section presents a comprehensive quantitative evaluation of NeRC's localization performance using real-world datasets. 

\subsection{Datasets}
\begin{figure}[!t]
  \centering
  \includegraphics[width=0.9\linewidth]{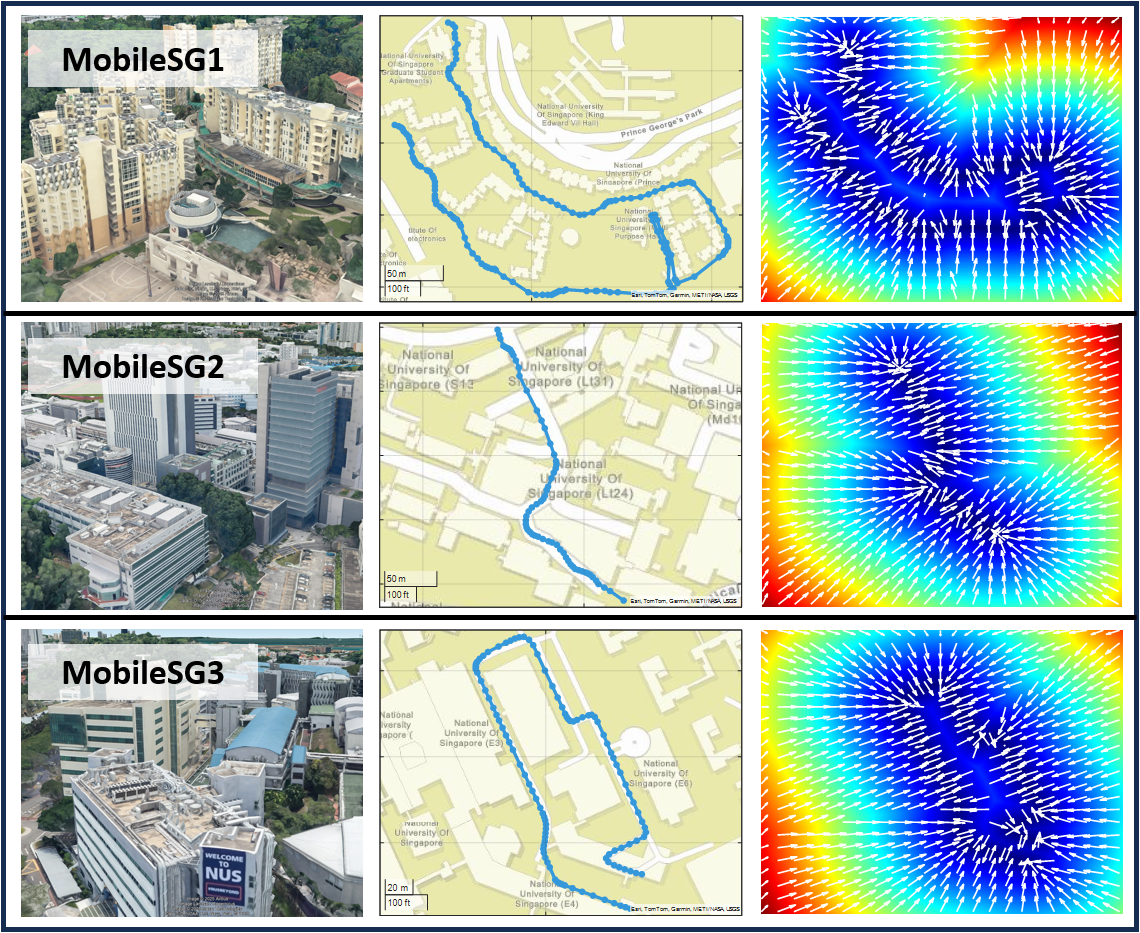}
  \caption{Three harsh urban scenes where we collected the MobileSG dataset to validate the EDF-based NeRC training.}
  \Description{This picture displays the environment where our MobileSG data was collected. The reference street data and corresponding EDF cost maps are also provided.}
  \label{fig: mobileSG}
\end{figure}

The GSDC datasets are publicly available \cite{fu2020android,fu2022summary} and provide a large volume of data collected {\bfseries repeatedly along consistent routes and within similar time intervals of days}, making them ideal benchmarks for evaluating NeRC’s performance. Accordingly, we leverage both the GSDC datasets and our own collected data (\autoref{fig: mobileSG}) to construct a range of evaluation scenarios. Unlike previous studies that relied solely on fully labeled datasets \cite{kanhere2022improving,10506762,xu2023differentiable,zhao2024improving}, {\bfseries our approach also exploits partially labeled data (with 2D annotations) and even unlabeled data}. A summary of the datasets used in our evaluation is presented in \autoref{tab:dataset}. 

\begin{table}[!t]
\caption{Dataset Description\label{tab:dataset}}
\centering
\begin{tabular}{c|cccccc}
\toprule
\multirow{2}{*}{Name} & {Urban} & {Train} & {Test}& {Length} & {Phone} & \multirow{2}{*}{Label}\\
&Level&Num.&Num.&(km)&Model&\\
\midrule
GSDC21R & Light&200k&32k&120&Pixel4&3D\\[0.8em]
GSDC21U & Harsh&37k&15k&10&Pixel4&3D\\[0.8em]
GSDC22-1 & Light&80k&41k&42&Pixel4&2D\\[0.8em]
GSDC22-2 & Light&116k&14k&55&Pixel4&2D\\[0.8em]
GSDC23 & Light&55k&9k&13&Pixel5&3D\\[0.8em]
MobileSG1 & Harsh&43k &6k &0.9& K40&EDF\\[0.8em]
MobileSG2 & Harsh&39k &2k&0.35&Mi9 &EDF\\
[0.8em]
MobileSG3 & Harsh&41k &2k &0.5&X50&EDF\\
\bottomrule
\end{tabular}
\end{table}

\begin{figure*}[!t]
  \centering
  \includegraphics[width=0.996\linewidth]{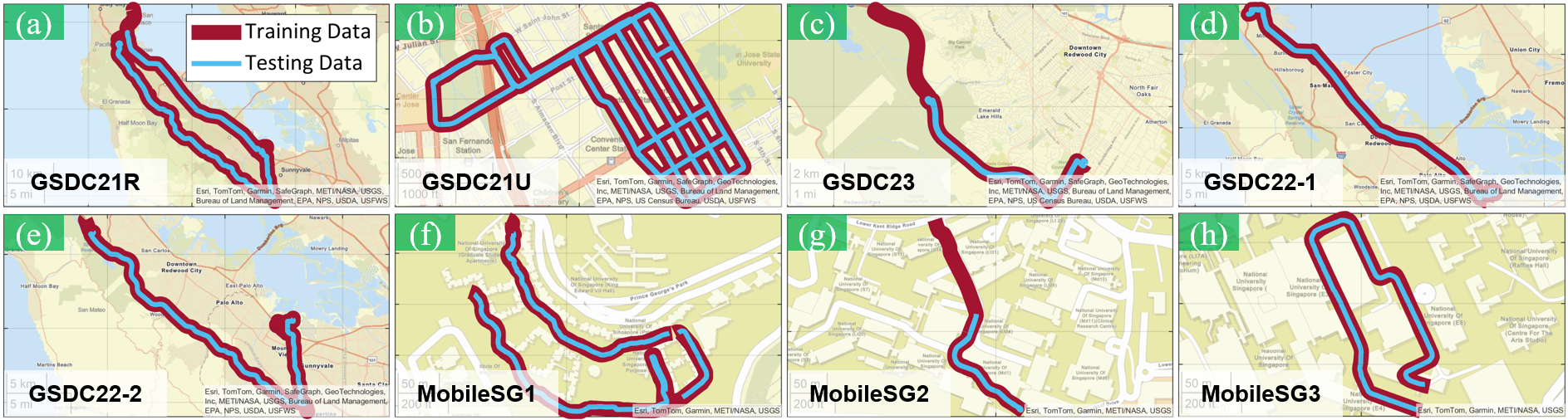}
  \caption{Visualization of datasets. (a) and (c)-(e) display open-sky rural areas. (b) and (f)-(h) show densely urban canyons.}
  \Description{The visualization of datasets. The training and testing data were collected along the same routes at approximately the same time intervals on different dates.}
  \label{fig:dataset}
\end{figure*}

{\bfseries GSDC 2021:} The dataset consists of raw GNSS measurements collected across the Mountain View and San Jose cities, covering both open and urban areas. Ground truth positions are fully labeled using the NovAtel SPAN integrated navigation system \cite{fu2020android}. We partition the dataset into two subsets: GSDC21R, containing data from rural areas, and GSDC21U, containing data from urban areas.


{\bfseries GSDC 2022:} Google enhanced the quality of their datasets in the second year of the competition, particularly in terms of ground truth calibration \cite{fu2022summary}. However, the dataset only contains trajectories in open areas, where ranging measurement errors are trivial. Additionally, most traces in this dataset are annotated with only 2D location labels. We extract from it two subsets along two trajectories: GSDC22-1 and GSDC22-2.


{\bfseries GSDC 2023:} The latest GSDC datasets offer similarly high-quality measurements as GSDC 2022, but with full 3D ground truth labels. From these, we extract a subset collected along a consistent route, referred to as GSDC23.

{\bfseries MobileSG:} We collected a dataset in extremely challenging urban canyon areas using various Android devices, as illustrated in \autoref{fig: mobileSG}. Ground truth locations were obtained only for the testing data, while the training data remained unlabeled. During testing data collection, the user’s movement was continuously recorded by a camera and later aligned with a 3D map in Google Earth Pro to determine the user's locations \cite{liu2018gnome}. This annotation method can achieve sub-meter accuracy \cite{sarlin2023orienternet}, which is sufficient given that positioning errors in urban environments can reach up to several hundred meters \cite{hsu2018analysis}. As shown in \autoref{fig: mobileSG}, the EDF cost maps are computed using publicly available street information corresponding to training trajectories. The MobileSG dataset is used exclusively for evaluating the performance of EDF-based training.

{\bfseries Summary:} The routes or scenes where all datasets were collected are visualized in \autoref{fig:dataset}, spanning a diverse set of collection routes, ranging from rural to urban environments and covering both large-scale and small-scale scenarios.

\subsection{Baseline Methods}
We compare the proposed NeRC framework against both traditional model-based methods and state-of-the-art data-driven approaches.

{\bfseries Model-based methods:} We choose the EKF, FGO (MHE-w/o-AC), and the full-information MHE (MHE-w/-AC) as baseline methods. To ensure a fair comparison, the covariance matrices and initial state are configured identically to those used in NeRC. 

{\bfseries Learning-based methods using intermediate loss:} FCNN-LSTM \cite{9490205} and PrNet \cite{10506762}, two SOTA deep learning-based pseudorange correction methods, are selected as baseline models. As they only predict ranging corrections without providing localization results, we integrate them with an MHE-based positioning engine---identical to the downstream component used in NeRC. Notably, they require 3D location labels to retroactively compute the ranging errors. As a result, they are inapplicable to partially labeled or unlabeled datasets such as GSDC22 and MobileSG.

{\bfseries End-to-end learning:} 
We also compared NeRC with SOTA end-to-end learning approaches, including the black-box GnssFormer and the structured learning paradigms E2E-PrNet \cite{10706359} and \verb|pyrtklib| \cite{hu2025pyrtklib,hernandez2024python}. GnssFormer is our enhanced adaptation of the Set Transformer \cite{kanhere2022improving}, incorporating the same input features used in NeRC and initialized with robust location estimates to guide its position corrections. Both E2E-PrNet and \verb|pyrtklib| adopt the structure composed of an MLP followed by a differentiable WLS positioning engine. E2E-PrNet also leverages the Theseus library for differentiable optimization, while \verb|pyrtklib| integrates a differentiable Python binding of the popular \verb|rtklib| library. We have enhanced these three baseline approaches with our proposed 2D and EDF-based training losses, enabling them to handle partially labeled or unlabeled data like NeRC. 

In a dataset, the same labels and loss function are used to train all data-driven methods. The baseline methods we implemented or trained are released at \url{https://github.com/AILocAR/NeRC}.




\begin{figure*}[!t]
\centering
\subfloat[GSDC21R]{\includegraphics[width=0.24\textwidth]{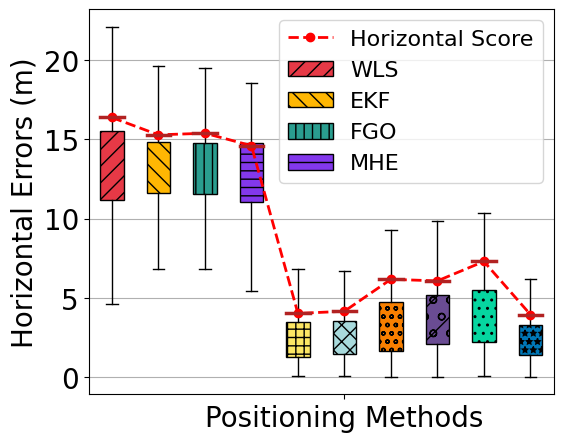}\label{fig:GSDC21R_results}
}
  \hfil
\subfloat[GSDC21U]{\includegraphics[width=0.24\textwidth]{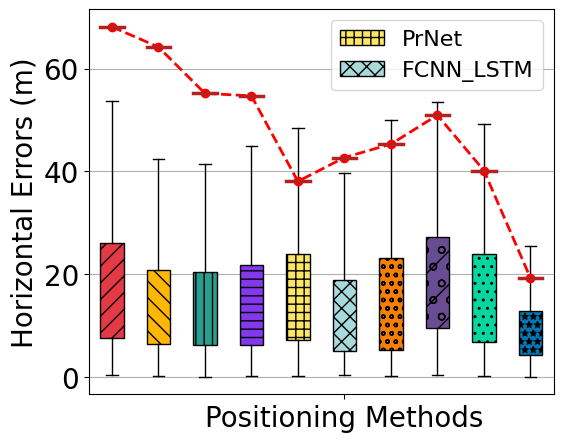}\label{fig:GSDC21U_results}}
\hfil
\subfloat[GSDC23]{\includegraphics[width=0.24\textwidth]{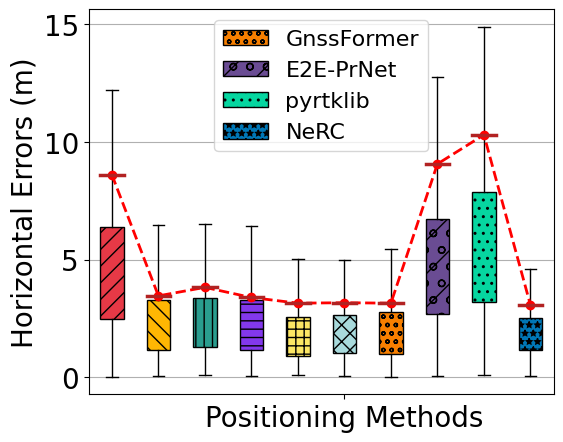}\label{fig:GSDC23_results}
}
  \hfil
\subfloat[GSDC22-1]{\includegraphics[width=0.24\textwidth]{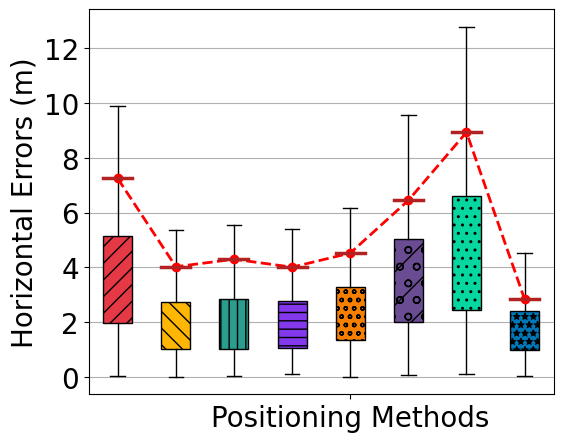}\label{fig:GSDC22_1_results}
}
  \hfil
\subfloat[GSDC22-2]{\includegraphics[width=0.24\textwidth]{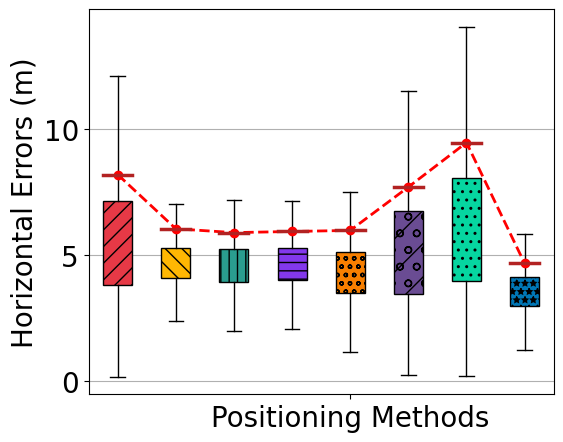}\label{fig:GSDC22-2_results}
}
\hfil
\subfloat[MobileSG1]{\includegraphics[width=0.24\textwidth]{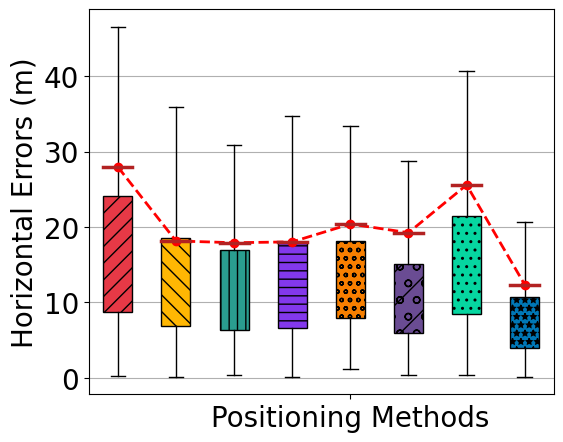}\label{fig:MobileSG1_results}}
\hfil
\subfloat[MobileSG2]{\includegraphics[width=0.24\textwidth]{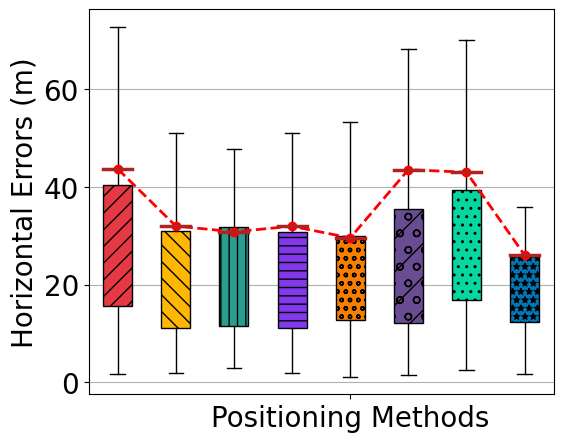}\label{fig:MobileSG2_results}
}
  \hfil
\subfloat[MobileSG3]{\includegraphics[width=0.24\textwidth]{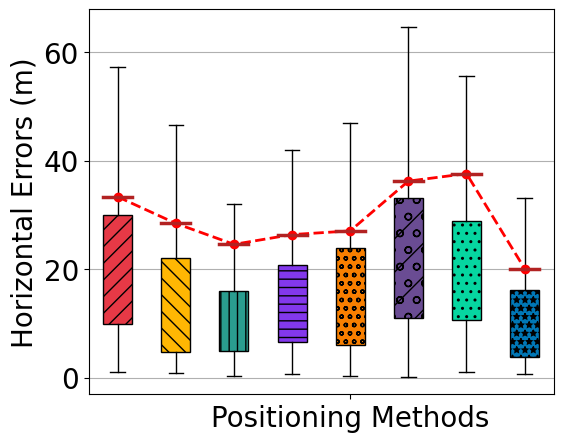}\label{fig:MobileSG3_results}
}
\caption{Horizontal positioning results.}
\Description{Horizontal positioning results.} 
\label{fig:benchmarking}
\end{figure*}

\begin{table*}\centering
  \caption{Horizontal Scores (Meter$\downarrow$) in Different Datasets}
  \label{tab:scores}
  \begin{tabular}{lll|cccc|cc|cccc}
    \toprule
   \multicolumn{3}{c|}{\multirow{2}{*}{Datasets}}&\multicolumn{4}{c|}{Model-based Methods}&\multicolumn{2}{c|}{Intermediate Loss}&\multicolumn{4}{c}{End-to-end Learning}\\
   \cline{4-13}
   &&&&&&&&&&&&\\[-0.8em]
   &&&WLS&EKF&FGO&MHE&PrNet&FCNN-LSTM&GnssFormer&E2E-PrNet&pyrtklib&NeRC\\
    \bottomrule
    \multicolumn{1}{c|}{}&\multicolumn{1}{c|}{}&&&&&&&&&&\\[-0.6em]
    \multicolumn{1}{c|}{\multirow{4}{*}{3D}}&\multicolumn{1}{c|}{\multirow{2}{*}{GSDC21}}&R &16.373 & 15.279 & 15.379 & 14.598  &4.019  &4.154&6.169&6.069&7.295 &\cellcolor{gray!25}{\textbf 3.948}\\
    \multicolumn{1}{c|}{}&\multicolumn{1}{c|}{}&&&&&&&&&&&\\[-0.8em]
    \multicolumn{1}{c|}{}&\multicolumn{1}{c|}{}&U&68.136 &64.201 & 55.190 &54.705 &38.071 &42.714&45.314 &50.959&40.180&\cellcolor{gray!25}{\textbf 19.301}\\
    \multicolumn{1}{c|}{}&\multicolumn{1}{c|}{}&&&&&&&&&&&\\[-0.8em]
    \cline{2-13} 
    \multicolumn{1}{c|}{}&\multicolumn{1}{c|}{}&&&&&&&&&&&
    \\[-0.6em]
    \multicolumn{1}{c|}{}&\multicolumn{1}{c|}{GSDC23}&\multicolumn{1}{c|}{-}&8.605 &3.460&3.821&3.391&3.150&3.169&3.149 &9.065&10.305&\cellcolor{gray!25}{\textbf 3.078}
    \\
    \multicolumn{1}{c|}{}&\multicolumn{1}{c|}{}&&&&&&&&&&&
    \\[-0.8em]
    \hhline{=============}
    \multicolumn{1}{c|}{}&\multicolumn{1}{c|}{}&&&&&&&&&&&\\[-0.6em]
    \multicolumn{1}{c|}{\multirow{2}{*}{2D}}&\multicolumn{1}{c|}{\multirow{2}{*}{GSDC22}}&1&7.248 &4.025 &4.299  &4.010  & \xmark & \xmark &4.522 &6.448 & 8.934&\cellcolor{gray!25}{\textbf 2.837}\\
    \multicolumn{1}{c|}{}&\multicolumn{1}{c|}{}&&&&&&&&&&&\\[-0.8em]
    \multicolumn{1}{c|}{}&\multicolumn{1}{c|}{}&2&8.204 &6.047  &5.896  &5.940 & \xmark & \xmark&5.983 &7.707&9.477&\cellcolor{gray!25}{\textbf 4.673}\\
    \multicolumn{1}{c|}{}&\multicolumn{1}{c|}{}&&&&&&&&&&&\\[-0.8em]
    \hhline{=============}
    \multicolumn{1}{c|}{\multirow{4}{*}{EDF}}&\multicolumn{1}{c|}{\multirow{4}{*}{MobileSG}}&&&&&&&&&&&\\[-0.8em]
    \multicolumn{1}{c|}{}&\multicolumn{1}{c|}{}&1 &27.957 &18.148 &17.909 &18.024 & \xmark & \xmark &20.387 &19.238 &25.572&\cellcolor{gray!25}{\textbf 12.350}\\
    \multicolumn{1}{c|}{}&\multicolumn{1}{c|}{}&&&&&&&&&&&\\[-0.8em]
    \multicolumn{1}{c|}{}&\multicolumn{1}{c|}{}&2 &43.687 &31.992   &30.824  &31.948  & \xmark & \xmark &29.570 &43.525 &43.002&\cellcolor{gray!25}{\textbf 26.043}\\
    \multicolumn{1}{c|}{}&\multicolumn{1}{c|}{}&&&&&&&&&&\\[-0.8em]
    \multicolumn{1}{c|}{}&\multicolumn{1}{c|}{}&3&33.332 &28.429  &24.568 &26.371  & \xmark & \xmark &27.050 &36.222 &37.602 &\cellcolor{gray!25}{\textbf 20.109}\\
  \bottomrule
\end{tabular}
\end{table*}

\subsection{Results}


{\bfseries Evaluation metrics:} Most location-based mobile applications require only horizontal positioning. In the GSDC competitions, evaluation is based solely on horizontal localization errors, measured by {\bfseries horizontal distances} between predicted and ground truth latitude/longitude coordinates. Accordingly, this work evaluates only horizontal positioning performance. The horizontal distance is calculated using Vincenty's formulae \cite{vincenty1975direct}, as implemented in the \verb|pymap3d| library. \autoref{fig:benchmarking} illustrates boxplots of horizontal errors of all the methods previously mentioned. Additionally, in accordance with the GSDC competition requirements, we report {\bfseries horizontal scores} for all methods, indicated by red dots in \autoref{fig:benchmarking} and summarized in \autoref{tab:scores}. The horizontal score is defined as the average of the 50th and 95th percentiles of horizontal errors. A lower horizontal score indicates higher positioning accuracy.



{\bfseries Positioning performance:}
\autoref{fig:benchmarking} and \autoref{tab:scores} indicate that NeRC outperforms all baseline methods across all datasets. 


(1) {\it Large-scale erroneous scenarios.} GSDC21R represents a large-scale erroneous scenario. As illustrated in \autoref{fig:dataset}a, the data were collected along 120 km of highways in the Bay Area, yet exhibit positioning errors of approximately 15 meters, as reflected in the model-based method results in \autoref{fig:GSDC21R_results}. Despite the open-sky environment, such inaccuracies may stem from system-level issues, particularly ground truth calibration errors \cite{fu2022summary}. In this case, all data-driven methods perform well, with NeRC achieving horizontal accuracy comparable to FCNN-LSTM and PrNet.



(2) {\it Large-scale open-sky scenarios.}
GSDC22 and GSDC23 were collected in large-scale open-sky scenes where model-based baselines already achieve high localization accuracy, owing to strong signal quality. In such environments, ranging errors are trivial and often comparable to inherent measurement noise. As a result, methods like E2E-PrNet and \verb|pyrtklib| tend to overfit to noise, leading to degraded performance. In contrast, NeRC, supported by differentiable MHE, effectively suppresses noise and corrects residual ranging errors. GnssFormer, learning location corrections in a black-box fashion, is sensitive to its initialization quality---here, EKF estimates---around which its outputs fluctuate \cite{mohanty2023learning, 10506762}. PrNet and FCNN-LSTM are trained on filtered ranging error labels \cite{10506762}, explaining their performance approaching NeRC in \autoref{fig:GSDC23_results}. However, their reliance on fully labeled 3D data precludes their use on GSDC22, where only 2D location labels are available.



(3) {\it Small-scale urban canyons.}
GSDC21U and MobileSG represent some of the most challenging small-scale urban localization scenarios. Despite severe multipath, NLOS effects, and high measurement noise, NeRC consistently outperforms all competing methods, achieving accuracy gains in both fully labeled and unlabeled settings. On the fully labeled GSDC21U dataset, NeRC surpasses the best model-based (MHE) and data-driven (PrNet) baselines by approximately 65\% and 49\%, respectively. On the unlabeled MobileSG dataset, NeRC improves over the leading model-based method (FGO) by 16–31\% and outperforms other end-to-end approaches (E2E-PrNet and GnssFormer) by 12–36\%. Therefore, NeRC reaches the SOTA level and maintains strong localization performance under these challenging conditions.





\begin{figure}[!t]
  \centering
\subfloat[]{
\includegraphics[width=0.475\linewidth]{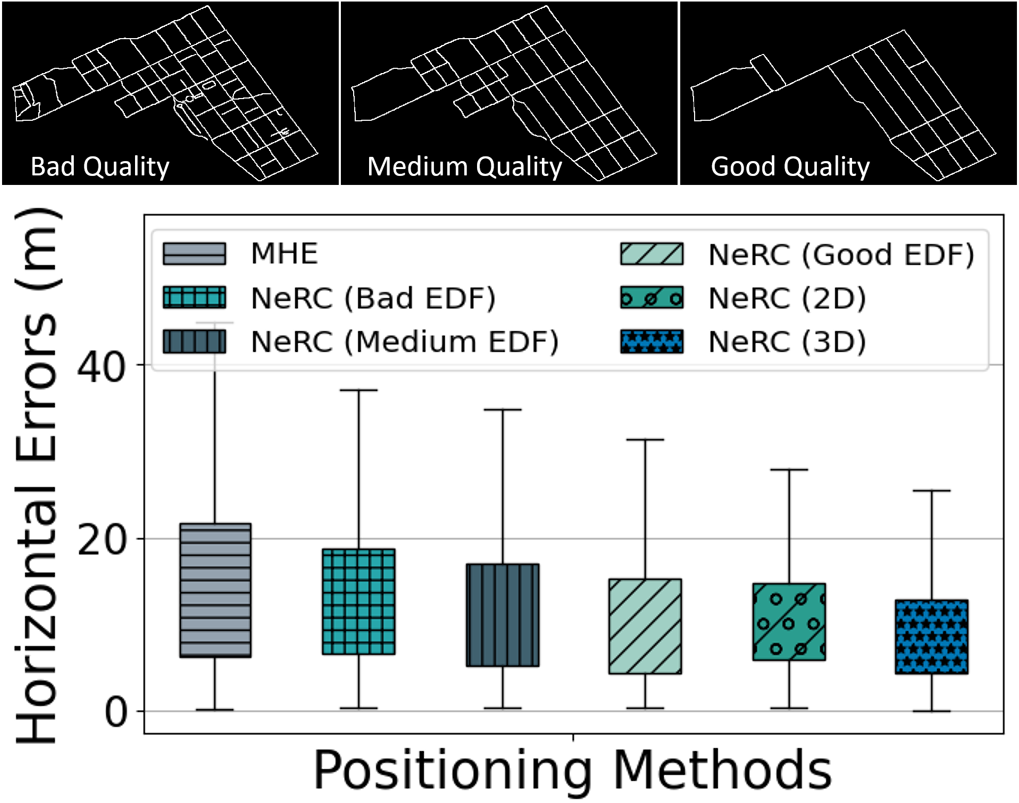}
\label{fig:training_methods}
}
\hfil
\subfloat[]{
\includegraphics[width=0.453\linewidth]{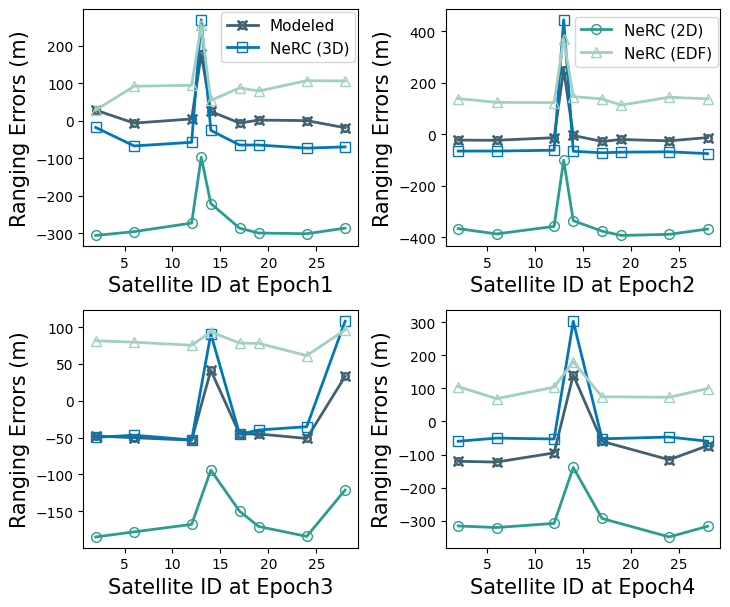}
\label{fig:ranging_error_prediction}
}
  \caption{(a) Horizontal errors of NeRC trained using different losses (b) Ranging errors predicted by the upstream MLP}
  \Description{These pictures show the impact of training loss and ranging errors regressed by the upstream MLP trained using 3D/2D ground truth labels and EDF cost maps.}
  \label{fig:impacts}
\end{figure}

{\bfseries Impact of training label quality:} Since NeRC is trained in a supervised manner, its performance is directly influenced by the accuracy of training labels. \autoref{fig:training_methods} compares the positioning errors of NeRC under different training strategies. The MHE algorithm serves as the baseline, and the same MHE is also integrated into NeRC as the downstream localization engine to demonstrate its enhancement. The evaluation is conducted on GSDC21U. Three EDF cost maps of varying qualities are provided for training with unlabeled data, reflecting different levels of prior knowledge about the routes used for data collection. As shown \autoref{fig:training_methods}, incorporating the EDF cost map can help reduce positioning errors without using location ground truth. Moreover, higher-quality EDF cost maps lead to smaller errors. Training with either 2D or 3D labels yields comparable performance, both contributing to further improvements in NeRC’s localization accuracy.



{\bfseries Ranging error regression:} \autoref{fig:ranging_error_prediction} displays ranging errors at four epochs regressed by the upstream MLP, which was trained using 3D/2D ground truth locations and EDF cost maps. The reference ranging errors are calculated based on 3D location labels, as described in \cite{10506762}. As illustrated in \autoref{fig:ranging_error_prediction}, at each epoch, the predicted ranging errors of visible satellites exhibit patterns similar to the reference, regardless of the training loss used. This indicates that the upstream MLP has successfully learned the underlying physical error patterns. For each kind of training loss, however, an offset nearly uniform across all satellites is observed between predictions and the reference, arising from missing label dimensions such as heights and device clock offsets \cite{10706359}. Nevertheless, the shared bias among all satellites only influences timing estimation and does not affect positioning accuracy \cite{kaplan2017understanding}.

\begin{figure}[!t]
  \centering
  \includegraphics[width=\linewidth]{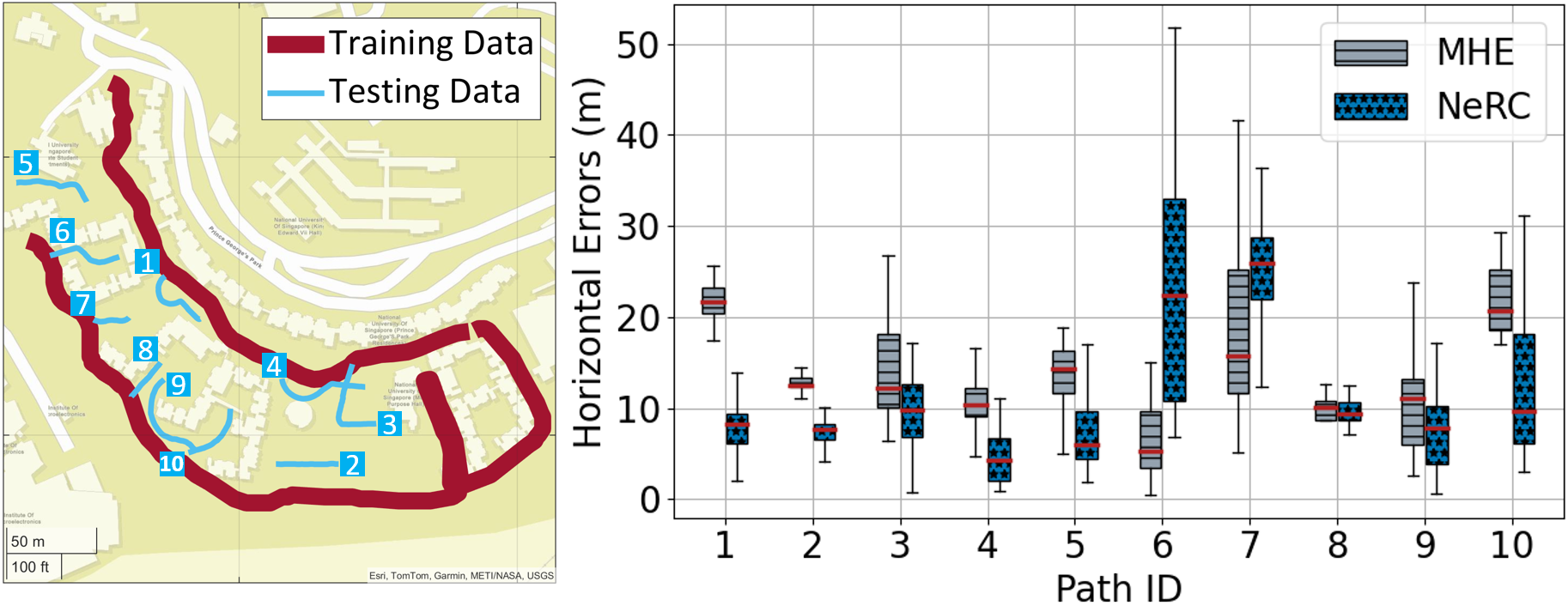}
  \caption{Generalization tests on unseen paths} 
  \Description{The testing paths are not covered by the training data.}
  \label{fig: generalization}
\end{figure}

{\bfseries Generalization ability:}
\autoref{fig: generalization} illustrates the robustness of NeRC in generalizing to unseen paths during inference. In this experiment, inference data were collected along ten alleyways that were not included in the training routes. Results show that NeRC can still enhance GNSS positioning accuracy in several unseen alleyways, even though the model was originally designed to learn within a specific route. However, on certain paths, such as Path 6 and Path 7, NeRC yields degraded positioning performance compared to the traditional model-based MHE. These paths are fully surrounded by unseen tall buildings, resulting in a distribution of ranging errors that differs significantly from the training data. 


\section{Real-Time Field Test}\label{sec:system_profiling}
One potential real-world application of NeRC is to provide ranging correction services over a network when users enter GNSS-challenged environments \cite{liu2018gnome,hernandez2024real}. To this end, we design an edge-based mobile localization system powered by NeRC and evaluate its real-time performance in urban canyon areas.



\begin{figure}[!b]
  \centering
  \includegraphics[width=\linewidth]{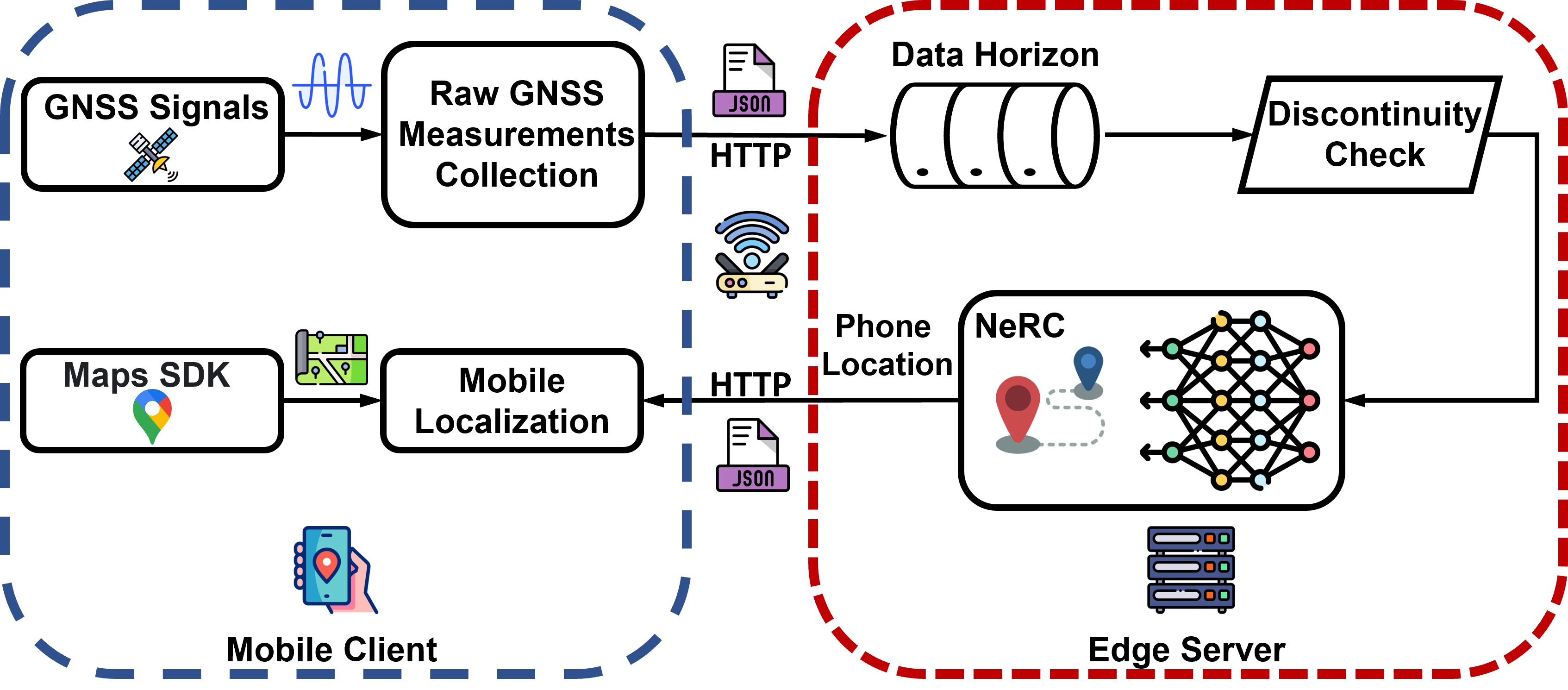}
  \caption{Edge-based mobile localization powered by NeRC} 
  \Description{The edge-based NeRC system for mobile localization is a typical client-server architecture.}
  \label{fig:nerc_ar}
\end{figure}

{\bfseries System Design:} \autoref{fig:nerc_ar} illustrates the diagram of the edge-based NeRC system for mobile positioning, which is a basic example of the client-server architecture. On the mobile client side, we developed an Android app to collect raw GNSS measurements and send them to the edge server via WLAN. Upon receiving data from smartphones, the edge server will cache it in a moving horizon, check for data discontinuities, and preprocess it for the subsequent neural ranging correction. Then, the server runs NeRC to compute user locations and send the results back to users via WLAN. By combining the map information accessed via the Google Maps API, we register user locations on 2D maps.

{\bfseries System Implementation:}
On the client side, we built the Android app in Java using Android Studio based on Google's open-source software GnssLogger \cite{fu2020android}. Raw GNSS measurements are collected with a frequency of 1 Hz via the Android Location API. Then, the collected data are packaged into JSON files for communicating with the edge server through the HTTP protocol \cite{scargill2023sitar}. The app runs on an Android 13 phone powered by the Snapdragon 8+ Gen 1 processor with 12 GB RAM. Only the GPS L1 signals are utilized in our experiment. The edge server is implemented using Uvicorn, an asynchronous Python web framework, to listen for and respond to requests from the Android client \cite{duan2024biguide}. The size of the horizon caching received data is set to 5. If the interval between the timestamps of the latest two messages exceeds 10 seconds, the data horizon will be considered discontinuous and reset to hold only the current message. NeRC is trained using EDF cost maps and just runs in inference mode \cite{hu2024seesys}. The server runs Ubuntu 18.04.6 LTS, powered by an Intel Core i7-10750H 2.6 GHz CPU with 16 GB RAM and an NVIDIA GeForce RTX 2060 GPU.


\begin{figure}[!t]
\centering
\subfloat[]{\includegraphics[width=0.409\linewidth]{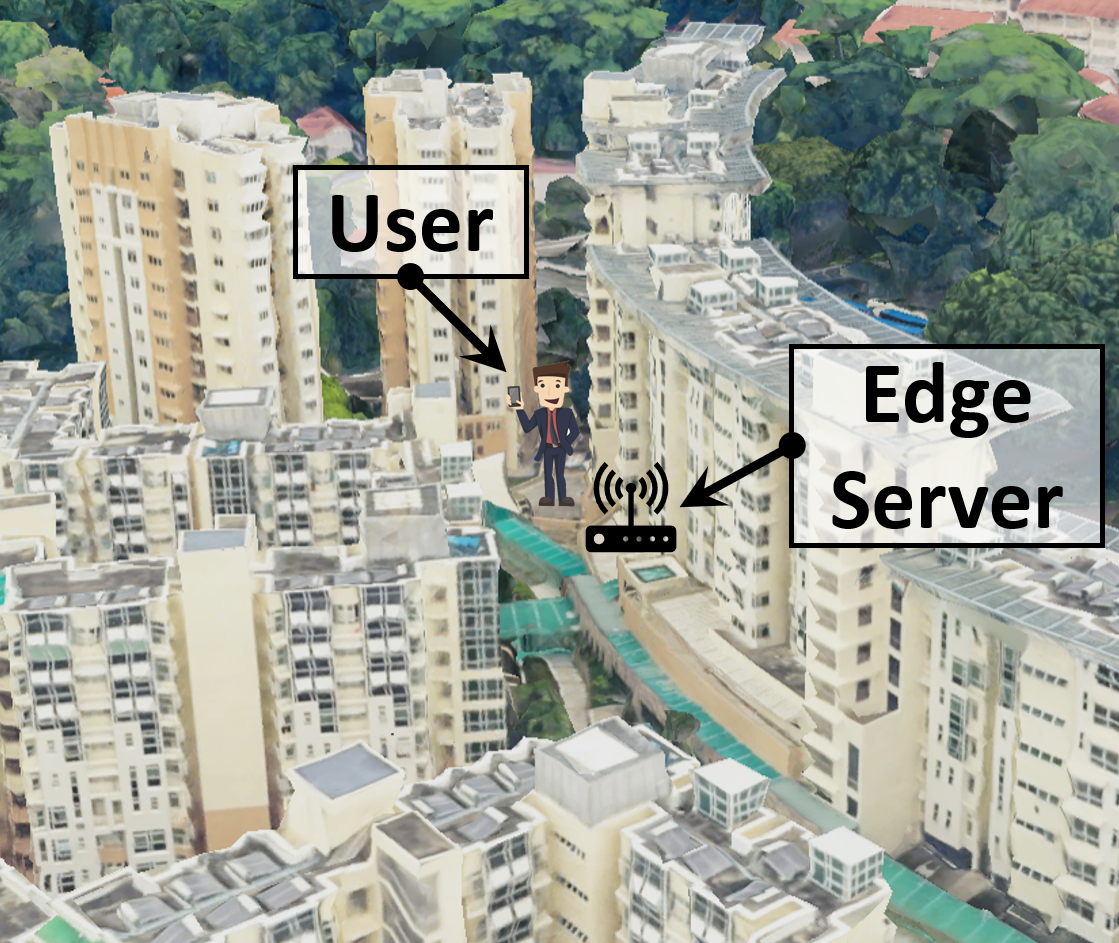}\label{fig:feild_test_scenario}
}
  \hfil
\subfloat[]{\includegraphics[width=0.259\linewidth]{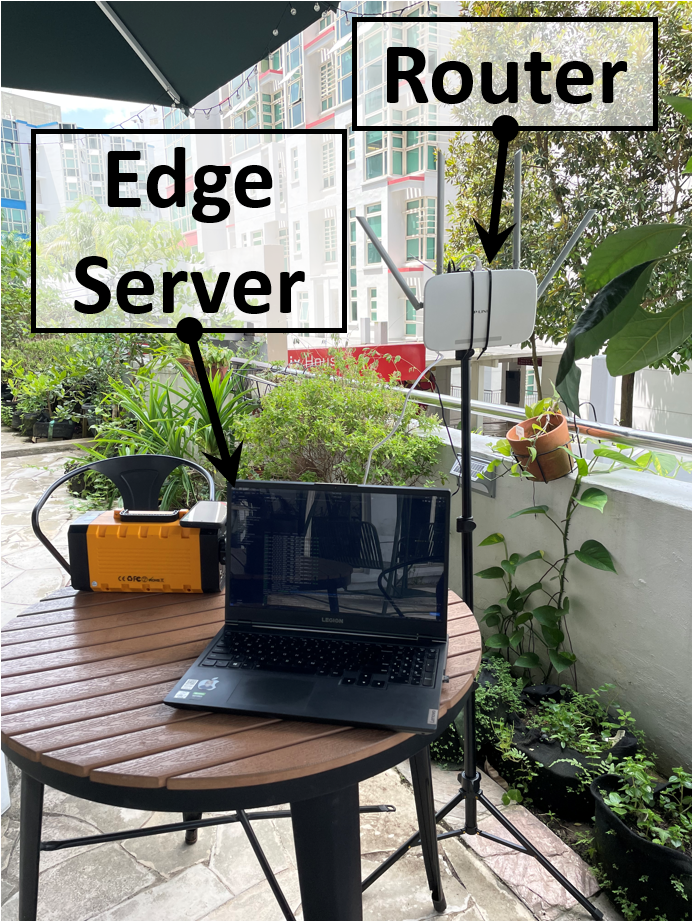}\label{fig: edge_server}}
\hfil
\subfloat[]{\includegraphics[width=0.258\linewidth]{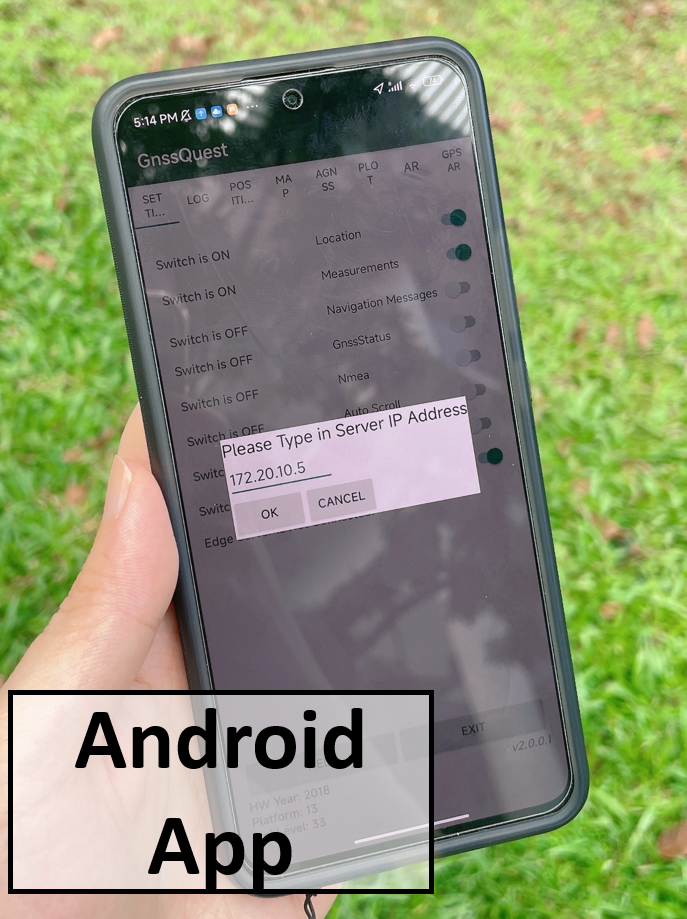}\label{fig:user_side}
}
\caption{(a) The experiment scene. (b) The edge server and its wireless access point. (c) Client Android application.}  
\Description{These figures depict the physical real-time field test, which is set by an edge server, a WAP, and a user mobile application. The experiment is conducted in a challenging scenario where multipath/NLOS conditions are severe.}
\label{fig:real_edge_Nerc}
\end{figure}

\autoref{fig:real_edge_Nerc} illustrates the physical setup of the real-time test. The experiment was conducted in a densely built urban area, where a road was surrounded by high-rise buildings, as shown in \autoref{fig:feild_test_scenario}. A user holding an Android phone walked along the road and localized himself using the NeRC system. As shown in \autoref{fig: edge_server}, a 2.4 GHz wireless access point was installed on a second-floor balcony to provide the widest possible coverage and to connect the server and the user within the same WLAN. Our Android app, displayed in \autoref{fig:user_side}, enables communication with the server once the user inputs the server’s IP address.

{\bfseries System Profiling:} This section focuses on quantitatively profiling the real-time performance of NeRC. Only a qualitative assessment of localization accuracy is provided here, as comprehensive evaluations on public and private benchmarks have already been presented in previous sections. 

\begin{figure}[!t]
\centering
\subfloat[]{\includegraphics[width=0.544\linewidth]{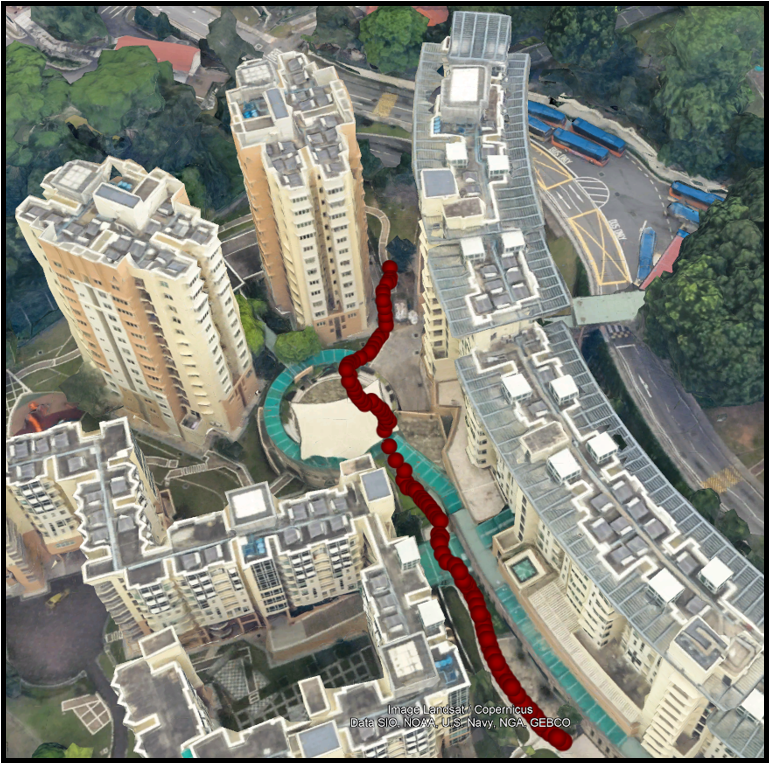}\label{fig:real_time_reference}
}
  \hfil
\subfloat[]{\includegraphics[width=0.117184\textwidth]{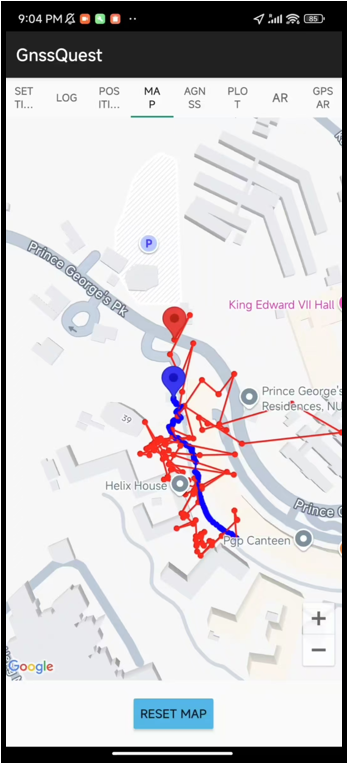}\label{fig: real_time_trajectory}}
\caption{Real-time test. (a) The reference trajectory. (b) The real-time localization trajectories: the red trajectory is Google's baseline solution, while the blue one is from NeRC.} 
\Description{These figures depict the physical real-time field test result. The experiment is conducted in a deep urban area. We walked along a pre-scheduled path and recorded the location trajectory from the mobile phone.}
\label{fig:real_test}
\end{figure}

(1) {\it Localization accuracy.} The reference trajectory is scheduled ahead of time, as shown in \autoref{fig:real_time_reference}. The user walked along the reference route, and the corresponding positioning result is presented in \autoref{fig: real_time_trajectory}. The red trajectory in \autoref{fig: real_time_trajectory} represents Google's baseline solution provided by the open-source GnssLogger software \cite{fu2020android}. Note that this is not the positioning result from the official Google Maps app, which utilizes in-phone GNSS chipset locations computed via the manufacturer’s proprietary algorithms. The baseline trajectory is visibly noisy and biased due to severe multipath/NLOS propagation effects from the surrounding high-rise buildings. In contrast, NeRC’s solution (depicted in blue) closely follows the reference trajectory, demonstrating higher accuracy and greater robustness to noise.

\begin{figure}[!b]
  \centering
  \includegraphics[width=0.96\linewidth]{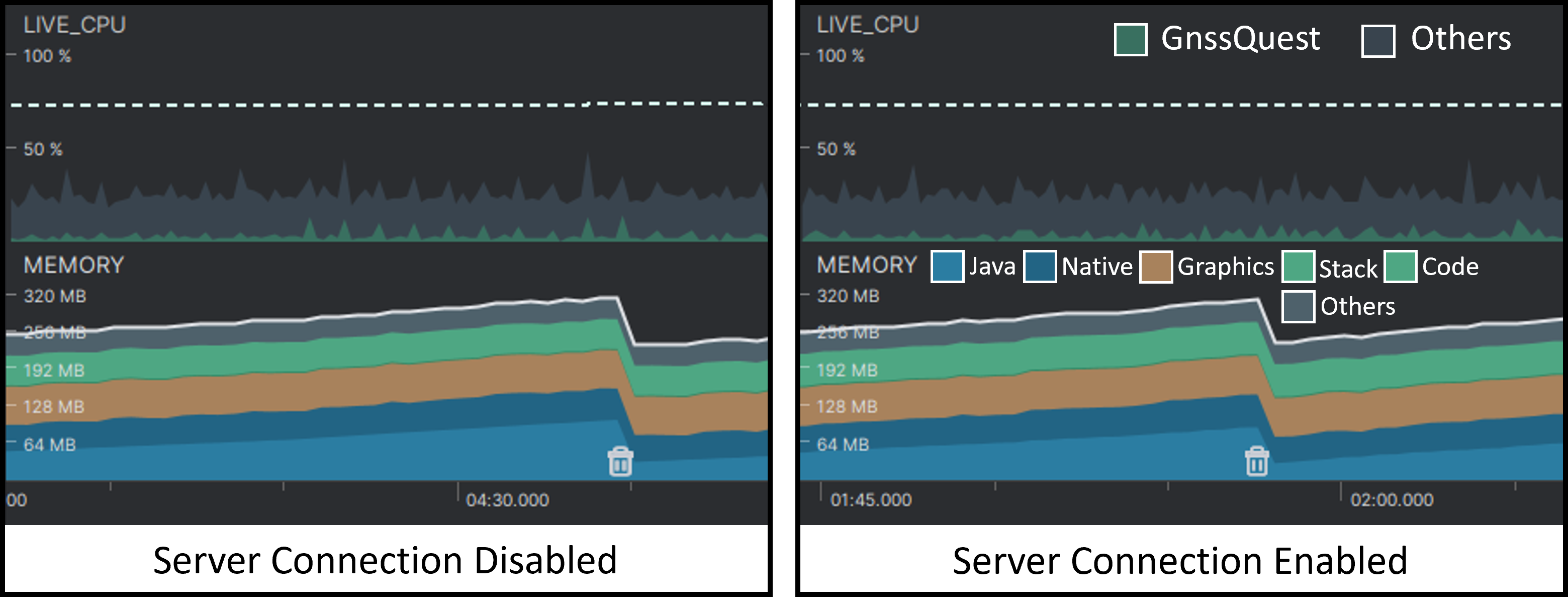}
  \caption{Phone CPU and memory usage} 
  \Description{The CPU and memory usage are profiled by ``View Live Telemetry" in Android Studio.}
  \label{fig: cpu_overhead}
\end{figure}

(2) {\it CPU overhead on mobile phones.} CPU and memory usage on the mobile device (Snapdragon 8+ Gen 1, 12 GB RAM) were measured using Android Studio’s live profiler \cite{xu2023practically,chen2024quantifying} under two conditions: with and without server connection. \autoref{fig: cpu_overhead} shows snapshots of the live profiling results. Our app, GnssQuest, demonstrates efficient performance in both cases. When the server is disconnected, average CPU usage is around 2\%, peaking at 13\%. The peak CPU usage is caused by map retrieval. Enabling the server connection adds only about 1\% additional CPU load. Mean memory usage is approximately 280 MB without the server, increasing by just around 10 MB when the connection is active.



(3) {\it GPU overhead on the edge server.} The edge server runs NeRC only in the inference mode. We trigger ``\verb|nvidia-smi|" programmatically to log the usage of the server's GPU and its memory per second. The average usage of GPU memory across 123 samples is 1.4 GB, while the free GPU memory is up to 4.5 GB. Moreover, the average utilization of the GPU is just 4.8\%. Thus, our NeRC model is lightweight enough to run on common consumer-level GPUs, such as the NVIDIA RTX 2060 we used.

(4) {\it Computational latency on the edge server.} We logged the inference time of NeRC on the edge server to evaluate its computational efficiency. On average, computing a location from a single epoch of measurements takes 100 ms, with a median latency of 98 ms. Inference time is closely influenced by the neural network size, the horizon length used in MHE, and the GPU's performance. Using a smaller neural network or horizon size on a more powerful GPU can reduce the latency.

(5) {\it End-to-end latency:} We define the end-to-end latency as the time interval required for the app to send raw measurements to the edge server and receive localization results in return \cite{duan2024biguide,ding2025rag}. Accordingly, we measured this latency on the client side. The end-to-end latency includes both the server-side computation time and the communication delay over the WLAN. In our experiment, the mean end-to-end latency was 246 ms, and the median was 225 ms. Given that our system operates at 1 Hz (i.e., Android collects raw GNSS measurements once per second), this level of latency has a negligible impact on the system’s real-time performance.



\section{Related Work}
{\bfseries Correcting Ranging Errors for GNSS:}
Correcting multipath and NLOS-induced pseudorange errors can enhance GNSS positioning, but existing solutions often require 3D city models \cite{miura2015gps,liu2018gnome} or additional sensors such as fisheye cameras or LiDAR \cite{wen2019correcting,bai2020using}, limiting widespread deployment. An alternative line of research uses deep learning to regress ranging errors directly from {\it \bfseries only} raw GNSS measurements. While promising, the related methods either depend on complex intermediate-loss estimation from fully labeled data \cite{sun2020improving,9490205,sun2023resilient,10506762}, making them label-intensive and less automatic. We address these limitations by designing a neural ranging correction framework that is trained and operates end-to-end.

{\bfseries Moving Horizon Estimation:} MHE is a well-established state estimation method effective for both linear \cite{muske1993receding,ling1999receding} or nonlinear \cite{rao2000nonlinear,rao2001constrained,rao2003constrained,weng2025recedinghorizonrecursivelocation} systems under the least-squares framework. It has been successfully applied in diverse domains, including chemical process control \cite{rao2002constrained,yan2025self}, agriculture \cite{li2024data}, satellite control \cite{zhao2025multi}, and robotics \cite{fiedler2020probabilistic, wang2021differentiable,10313083,kang2024fast}. More recently, its use in satellite navigation has also been explored \cite{wang2015robust,liu2023state,weng2023localization,liu2024actualization,weng2025recedinghorizonrecursivelocation}. Meanwhile, as an equivalent version of MHE under the assumption of Gaussian system statistics \cite{weng2025recedinghorizonrecursivelocation}, FGO has also demonstrated outstanding performance in positioning using raw GNSS measurements \cite{suzuki2023precise, wen2021factor,xu2023differentiable,xu2024autow,liu2025attitude}. However, prior studies have not examined the performance of MHE when integrated into a learning pipeline, particularly regarding its forward and backward performance. Our work addresses this gap.

{\bfseries End-to-end Structured Learning:} The learning paradigm that integrates data-driven neural modules and differentiable physical reasoning has been widely adopted in robotics \cite{li2020towards,wang2021differentiable,10313083,cao2024differential}, autonomous driving \cite{huang2023differentiable,liu2025hybrid}, 3D vision \cite{chen2022epro,valassakis2024handdgp}, optical \cite{10.1007/978-3-030-58452-8_24} and radio frequency reconstruction \cite{zhao2023nerf2,zhao2024crowdsourced}, as well as other domains requiring physical or logical priors \cite{pineda2022theseus}. This paradigm has also been formalized under the high-level concept of {\it \bfseries Imperative Learning} \cite{wang2025IL,Yang-RSS-23,zhan2024imatching,fu2024islam}. Recent studies have brought this idea to GNSS localization, including end-to-end learning-based LOS/NLOS signal weighting \cite{xu2023differentiable,xu2024autow} and ranging error correction \cite{10706359,hu2025pyrtklib}. Regarding ranging error regression, existing methods have so far relied solely on a baseline WLS localization engine and have been trained using fully labeled data \cite{10706359,hu2025pyrtklib}. This paper addresses both limitations by introducing a differentiable, noise-resilient location estimator and incorporating basic 2D public maps to enable training with ubiquitous unlabeled GNSS data.





\section{Conclusion, Limitations, and Future Work}
This paper presents NeRC, a novel neural ranging correction framework that enhances GNSS localization for mobile devices. NeRC is trained through differentiable MHE, guided by location-based losses to ensure end-to-end optimization. Based on this, we propose a new training paradigm that can leverage unlabeled GNSS data by incorporating Euclidean-distance supervision derived from publicly available maps. Extensive evaluations across both public and private benchmarks demonstrate the robustness and superiority of NeRC across rural
and urban environments under both labeled and unlabeled settings. Furthermore, an edge-based field test reveals its promising potential for practical deployment. The following limitations of NeRC merit further investigation:




(1) {From centralized to distributed}: Enabling onboard training and inference with data from heterogeneous devices is important for NeRC's distributed deployment \cite{du2023accelerating,xu2023practically,du2025decentralized}. It is also necessary to systematically investigate the spatial-temporal granularity at which NeRC models should be trained and distributed for large-scale use.



%

(2) {From deep learning to pretraining}: NeRC learns from unlabeled data under the assumption that we know data collection routes. Exploiting crowdsourced data with minimal prior knowledge could pave the way for GNSS foundation models \cite{smith2024mapping}.

(3) {From fingerprinting to generalization}: NeRC is currently designed to regress ranging errors of satellites observed in the same region during similar periods of days. Incorporating other sensing modalities shows potential for generalizable error modeling \cite{weng2024gnssquest}.



\begin{acks}
This work was supported by \grantsponsor{200604393R}{Nanyang Technological University}{https://www.ntu.edu.sg/} under the NTU Research Scholarship:˜\grantnum[https://www.ntu.edu.sg/admissions/graduate/financialmatters/scholarships/rss]{200604393R}{R2015378}. Kun Cao was supported by the National Natural Science Foundation of China under Grant 62088101, 62503367, Shanghai Municipal Science and Technology Major Project (No. 2021SHZDZX0100).
\end{acks}

\bibliographystyle{ACM-Reference-Format}
\bibliography{main}


\end{document}